\documentclass[9pt,technote,twoside,twocolumn]{IEEEtran}
\ifCLASSOPTIONcompsoc
  \usepackage[nocompress]{cite}
\else
\fi

\ifCLASSINFOpdf
   \usepackage[pdftex]{graphicx}
   \DeclareGraphicsExtensions{.pdf,.jpeg,.png}
\else
\fi

\usepackage[cmex10]{amsmath}
\usepackage{times}
\usepackage{graphicx,epstopdf}
\usepackage{amsmath,eucal,amssymb}
\usepackage{amssymb,bibspacing,url,xspace}
\usepackage[bf,small]{caption}

\usepackage{mdwtab}
\usepackage{multirow}

\usepackage[small]{subfigure}

\usepackage[]{cite}

\usepackage{algorithm2e}
\let\savedalgorithm\algorithm
\let\savedendalgorithm\endalgorithm

\usepackage{algorithm}

\newcommand{\st}{{\rm s.t.}\xspace}

\def\argmax{\operatorname*{argmax\,}}

\def\T{{\!\top}}

\def\Real{\mathbb{R}}

\newcommand{\fnorm}[2][2]{\ensuremath{ \left\| #2 \right\|_{ \mathrm{#1} } } }

\newcommand{\iprod}[1]{\ensuremath{ \left< #1 \right> } }

\newtheorem{result}{Claim}[section]

\def\bxi{ {\boldsymbol{\xi} } }

\def\bx{ {\bf x } }

\def\by{ {\bf y } }

\def\bb{ {\bf b } }
\def\bw{ {\bf w } }
\def\bs{ {\bf s } }

\def\bc{ {\bf c } }

\def\bS{ {\bf S } }
\def\bI{ {\bf I } }

\def\bK{ {\bf K } }
\def\bL{ {\bf L } }
\def\bO{ {\bf O } }

\def\bU{ {\bf U } }

\def\calX{ {\cal X } }
\def\calF{ {\cal F } }
\def\calH{ {\cal H } }
\def\calW{ {\cal W } }

\def\st{ {\rm s.t. } }
\def\eg{{\rm e.g. }}
\def\ie{{\rm i.e. }}

\def\etc{{\rm etc.}}

\def\fsv{ f_{\rm sv} }
\def\flp{ f_{\rm lp} }

\def\spx{ {\rm spx }}

\def\liblinear{{\sc liblinear}\xspace}

\def\CGENS{{\sc  CGEns}\xspace}

\def\lssvm{{\sc  CGEns-SLS}\xspace}

\def\half{{\tfrac{1}{2} }}
\def\ones{{\boldsymbol  1}}

\def\balpha{{\boldsymbol  \alpha}}

\def\wl{{\hslash  }}

\def\H{{\bf H}}
\def\zeros{{\boldsymbol  0}}
\def\bh{{\bf h}}

\def\sign{{\rm sign}}

\begin{document}

\title{From Kernel Machines to Ensemble Learning}

\author{
         Chunhua Shen,
         Fayao Liu
\thanks
{
}
\thanks
{
The authors are with Australian Center for Visual Technologies,
and School of Computer Science,
The University of Adelaide, SA 5005, Australia
(e-mail: \{chunhua.shen,fayao.liu\}@adelaide.edu.au).
}
\thanks
 {
 This work was in part supported by Australian Research Council Future
 Fellowship FT120100969.
 }
}

\maketitle

\begin{abstract}

        Ensemble methods such as boosting combine multiple learners to obtain better prediction than
        could be obtained from any individual learner.  Here we propose a principled framework for
        {\em directly} constructing ensemble learning methods from kernel methods.  Unlike previous
        studies showing the equivalence between boosting and support vector machines (SVMs), which
        needs a translation procedure, we show that it is possible to design boosting-like
        procedure to solve the SVM optimization problems.
        In other words, it is possible to design ensemble methods directly from SVM
        without any middle procedure.
        This finding not only enables us to design new ensemble learning methods directly from kernel
        methods, but also makes it possible
        to take advantage of those highly-optimized fast linear SVM solvers for ensemble learning.
        We exemplify this framework for designing binary ensemble learning as well as a new
        multi-class ensemble learning methods.
        Experimental results demonstrate the flexibility and usefulness of the proposed framework.

\end{abstract}


\section{Introduction}

        Ensemble learning methods, with a typical example being boosting
        \cite{freund1997,Schapire98,Dual2010Shen,Paisitkriangkrai2013RandomBoost,MDBoost2010Shen}, have
        been successfully applied to many machine learning and computer vision applications.
        Its excellent performance and fast evaluation have made ensemble learning one of
        the most widely-used learning methods, together with kernel machines like SVMs.
        In the literature, the general connection between boosting and SVM has been shown
        by Schapire et al. \cite{Schapire98}, and R\"atsch et al.\ \cite{Ratsch02}.
        In particular, R\"atsch et al.\ \cite{Ratsch02} developed a mechanism to convert SVM
        algorithms to boosting-like algorithms by translating the quadratic programs (QP) of SVMs into
        linear programs (LP) of boosting (similar to LPBoost \cite{LPB}).
        A one-class boosting method was then designed by converting one-class SVM into an LP.
        Following this vein, a direct approach to multi-class boosting was developed in
        \cite{multiboost} by using the loss function in Crammer and Singer's multi-class SVM
        \cite{multisvm01}.
        The recipe to transfer algorithms is essentially \cite{Ratsch02}:
        ``The SV-kernel is replaced by an appropriately constructed hypothesis space for leveraging
        where the optimization of {\it an analogous mathematical program is done using $ \ell_1$
        instead of $ \ell_2$-norm.}''
        This transfer is {\it indirect} in the sense that one has to design a different $ \ell_1$
        norm regularized mathematical program.
        We suspect that this is due to the widely-adopted belief that boosting methods need the
        sparsity-inducing $ \ell_1$-norm regularization so that the final ensemble model only relies
        on a subset of weak learners \cite{Ratsch02,Dual2010Shen}.\footnote{At the same time,
        standard SVM needs $ \ell_2 $ regularization so that the kernel trick can be applied,
        although $ \ell_1 $ SVM \cite{Zhu03norm1} takes a different approach.
        }
        In this work, we show that it is possible to design ensemble learning methods by
        {\it directly} solving standard SVM optimization problems.
        Unlike \cite{Ratsch02,multiboost}, no mathematical transform is needed.
        The only optimization technique that our framework relies on is column generation.
        With the proposed framework, the advantages that we can think of are:
        1) Many kernel methods can directly have an equivalent ensemble model;
        2) As conventional boosting methods, our ensemble models are iteratively learned. At each
        iteration, compared with the $ \ell_1 $ optimization
        involved in the indirect approach \cite{Ratsch02,LPB,Dual2010Shen,multiboost}, our
        optimization problems are much simpler.  For the first time, we enable the use of
        fast linear SVM optimization software for ensemble learning.
        3) As the fully-corrective boosting methods in \cite{Dual2010Shen,multiboost}, our ensemble
        learning procedure is also fully corrective. Therefore the convergence speed is often much
        faster than stage-wise boosting.
        4) Kernel SVMs usually offer promising classification accuracies at the price of high
        usage of memory and evaluation time, especially when the size of training data is large.
        Recall that the number of support vectors is linearly proportional to the number of training
        data \cite{Steinwart2003}.
        Ensemble models, on the other hand, are often much faster to evaluate.
        Ensemble learning is also more flexible in that the user can determine the number of weak
        learners used. Typically an ensemble model uses less than a few thousand weak learners.
        Ensemble learning can also select features by using decision stumps or trees as weak
        learners, while nonlinear kernels are defined on the entire feature space.
        The developed framework tries to enjoy the best of both worlds of kernel machines and
        ensemble learning.
        Additional contributions of this work include:
        1) To exemplify the usefulness of this proposed framework, we introduce a new multi-class
        boosting method based on the recent multi-class SVM \cite{simplex12}.
        The new multi-class boosting is effective in performance and can be efficiently learned
        since a closed-form solution exists at each iteration.
        2) We introduce Fourier features as weak learners for learning the strong classifier.
        Fourier features approximate the radial basis function (RBF) Gaussian kernel.
        Our experiments demonstrate that Fourier weak learners usually outperforms decision stumps
        and linear perceptrons.
        3) We also show that multiple kernel learning is made much easier with the proposed
        framework.

        \section{Related work}
        The general connection between SVM and boosting has been discussed by
        a few researchers \cite{Schapire98,Ratsch02} at a high level.  To our knowledge, the work
        here is the first one that attempts to build ensemble
        models by solving SVM's optimization  problem. We review some closest work next.
        Boosting has been extensively studied in the past decade
        \cite{freund1997,Schapire98,multiboost,LPB,Dual2010Shen}. Our methods are close to
        \cite{LPB,Dual2010Shen} in that we also use column generation (CG) to select weak learners and
        fully-correctively update weak learners' coefficients.  Because we are solving the SVM
        problem, instead of the $ \ell_1 $ regularized boosting problem,
        conventional CG cannot be directly applied. We use CG in a novel way---instead of looking at
        dual constraints, we rely on the KKT conditions.

        If one uses an infinitely many  weak learners in boosting \cite{Lin2008}
        (or hidden units in neural
        networks \cite{CNN07}), the model is equivalent to SVM with a certain kernel.
        In particular,  it shows that when the feature mapping function $ \Phi(\cdot)$ contains
        infinitely many randomly distributed decision stumps, the kernel function $ k( \bx, \bx' ) = \iprod{ \Phi(\bx),
        \Phi(\bx')  } $ is the stump kernel of the form $ \Delta - \Vert \bx - \bx ' \Vert_1 $. Here
        $ \Delta $ is a constant, which has no impact on the SVM training.
        Moreover,
        when $ \Phi(\bx) = {\rm sign} ( \theta ^\T \bx  - \kappa ) $, \ie, a perceptron,
        the corresponding kernel is called the perceptron kernel
        $ k( \bx, \bx' ) = $ $ \Delta' - \Vert \bx - \bx ' \Vert_2 $.

        Loosely speaking, boosting can be seen as explicitly computing the kernel
        mapping functions because, as pointed out in \cite{Ratsch02}, a kernel constructed by the
        inner product of weak learners' outputs satisfies the Mercer's condition.
        Random Fourier features (RFF) \cite{Rahimi07} have been applied to large-scale kernel methods.
        RFF is designed by using the fact that a shift-invariant kernel is the Fourier transform of
        a non-negative measure. Yang et al.\ show that RFF does not perform well due to its
        {\em data-independent}  sampling strategy when there is a large gap in the eigen-spectrum of
        the kernel matrix \cite{Yang2012NIPS}.
        In \cite{MajiTpami2012,vedaldi12},  it shows that for homogeneous additive kernels, the kernel
        mapping function can be exactly computed.
        When RFF is used as weak learners in the proposed framework here, the greedy CG based RFF
        selection can be viewed as {\em data-dependent} feature selection. Indeed,  our
        experiments demonstrate that our method performs much better than random sampling.

        \section{Kernel methods and ensemble models}
        We first review some fundamental concepts in SVMs and boosting.
        We then show the connection between these two methods and show how to design column
        generation based ensemble learning methods that directly solve the optimization problems in
        kernel methods.

        Let us consider binary classification for the time being. Assume that the input data points are
        $
        ( \bx_i, y_i ) $ $ \in \calX $ $ \times \{ -1, 1 \},
        $ with $ i = 1\cdots m. $
        For SVMs, it is well known that the original data $ \bx $ are implicitly mapped to a feature
        space through a mapping function $ \Phi: \calX \rightarrow \calF $. The function $ \Phi $ is
        implicitly defined by a kernel function $  k ( \bx, \bx' ) = \iprod{ \Phi( \bx ), \Phi(\bx')} $, which
        computes the inner product in $ \calF $.
        SVM finds a hyperplane that best separates the data by solving:
        \begin{align}
            \label{EQ:SVM1}
            \min_{ \bw, b, \bxi \geq 0 }
                    \fsv = \half \fnorm{\bw}^2 + C \ones^\T \bxi,
        \end{align}
        subject to the margin constraints $   y_i ( \bw^\T \Phi(\bx_i) + b ) \geq 1 - \xi_i $,
        $ \forall i$. Here $ \ones $ is a vector of all $ 1$'s.
        The Lagrange dual can be easily derived:
        \begin{align}
            \label{EQ:SVM_D1}
            \max_{ \balpha }  \ones ^\T \balpha - \half \balpha  ^\T  ( \bK \circ   \by \by^\T )
            \balpha, \st\,
            0 \leq \balpha \leq C, \by ^\T \balpha = 0.
        \end{align}
        Here $ C $ is the trade-off parameter; $ \bK $ is the kernel matrix with
        $ \bK_{ij} = k( \bx_i,
        \bx_j)$; and $ \circ $ denotes element-wise matrix multiplication, \ie, Hadamard product.
        $\by \by^\T $ is the label matrix with  $ \by = [y_1, $ $ \cdots, $ $ y_m ]^\T $.
        Note that in the case of linear SVMs, \ie, $ k(\bx, \bx') = \iprod{ \bx, \bx'  } $,
        there  are fast and scalable algorithms for training linear SVMs, \eg, \liblinear
        \cite{liblinear}.

        Ensemble learning methods, with boosting being the typical example, usually learns a strong
        classifier/model by linearly combining a finite set of weak learners. Formally the learned
        model is $ F(\bx) =  \bw^\T  \Phi( \bx) $ with
        \begin{align}
            \label{EQ:PhiBoost}
            \Phi(\bx)
            =  [ \wl_1( \bx ),   \cdots,  \wl_J(\bx)  ]^\T.
        \end{align}
        Therefore, in the case of boosting,
        the feature mapping function is explicitly learned:
        $
            \Phi: $ $\bx \rightarrow $ $ [  \wl_1( \bx ), $ $ \cdots,  $ $ \wl_J(\bx)  ]^\T,
        $
        where $ \wl ( \cdot )  \in \calH $ is the weak learner.
        It is easy to see that a kernel induced by the weak learner set
        $ k (\bx, \bx' ) = \sum_j \wl_j ( \bx ) \wl_j( \bx' ) $ is a valid one and its corresponding
        kernel matrix must be positive semidefinite.
        Next let us take LPBoost as an example to see how CG is used to explicitly
        learn weak learners, which is the core of most boosting methods \cite{LPB,Dual2010Shen}.

        The primal program of LPBoost can be written as
        \begin{equation}
            \label{EQ:LPB1}
             \min_{ \bw \geq 0,  \bxi \geq 0 }
             \flp = \fnorm[1]{\bw} + C \ones^\T \bxi,
        \end{equation}
        subject to the margin constraint $  y_i ( \bw^\T \Phi(\bx_i) ) \geq 1 - \xi_i  $,
        $ \forall  i$.
        with $ \Phi(\cdot) $ defined in \eqref{EQ:PhiBoost}.
        The dual of \eqref{EQ:LPB1} is
        \begin{align}
            \label{EQ:LPB_Dual}
            \max_\balpha \,
                \ones^\T \balpha, \, \st\,
                0 \leq \balpha \leq C, {\textstyle \sum}_i y_i \alpha_i \Phi( \bx_i ) \leq \ones.
        \end{align}
        Note that the last constraint in the dual is a set of $ J $ constraints.
        Often, the number of possible weak learners can be infinitely large. In this case it is
        intractable to solve either the primal or dual.
        In this case, CG can be used to solve the problem.
        These original problems are referred to as the master problems. The
        CG method solves these problems by incrementally selecting a subset
        of columns (variables in the primal and constraints in the dual) and optimizing the
        restricted problem on the subset of variables.
        So the basic idea of CG  is to add one constraint at a time to the dual problem
        until an optimal solution is identified.
        In terms of the primal problem,
        CG  solves the problem on a
        subset of variables, which corresponds to a subset of constraints in the dual.
        If a constraint absent from the dual problem is violated by the solution to the restricted
        problem, this constraint needs to be included in the dual problem to further restrict its
        feasible region.
        To speed convergence we
        would like to find the one with maximum deviation (most violated dual constraint),
        that is, the base learning algorithm
        must deliver a function $ \hat \wl ( \cdot ) $ such that
        \begin{align}
            \label{EQ:WL}
            \hat \wl ( \cdot )  = \argmax_{ \wl \in \calH} \; \textstyle \sum_i y_i \alpha_i
            \wl( \bx_i ).
        \end{align}
        If there is no weak learner
        $ \wl    (\cdot) $ for which the dual constraint
        $ \sum_i y_i \alpha_i  \wl( \bx_i ) \leq 1 $ is violated, then the current combined
        hypothesis is the optimal solution over all linear combinations of weak learners.
        That is the main idea of LPBoost \cite{LPB} and its extension \cite{Dual2010Shen}.
        It has been believed that here two components have played an essential role
        in this procedure of deriving this meaningful dual such that CG can be applied. 1) This
        derivation relies on the $ \ell_1 $ norm regularization in the primal objective of $ \flp $.
        2) The constraint of nonnegative $ \bw $ lead to the dual inequality constraint.
        Without this nonnegative constraint, the last dual constraint becomes an equality:
        $ {\textstyle \sum}_i y_i \alpha_i \Phi( \bx_i ) = \ones $.  In terms of optimization,
        the constraint $  \bw \geq 0  $ causes difficulties.
        We will show the remedies for these difficulties in the next section.

        \section{From SVM to ensemble learning}
        We show how to derive ensemble learning directly from kernel methods like SVM.
        Our goal is to explicitly solve \eqref{EQ:SVM1}
        {\it without using the kernel trick}. In other words, similar to boosting, we iteratively
        solve \eqref{EQ:SVM1} by {\it explicitly learning the kernel mapping function $ \Phi(\cdot)$}.
        At the first glance, it is unclear how to use the idea of CG to derive a boosting-like
        procedure similar to LPBoost, as discussed above.
        In order to add a weak learner $ \wl(\cdot) $ into $ \Phi(\cdot) $
        by finding the most violated dual constraint---as a starting point---we must
        have a dual constraint containing
        $ \Phi(\cdot) $.
        From the  dual problem of SVM \eqref{EQ:SVM_D1}, the main difficulty here is
        that the dual constraints are two types of simple linear constraints on the dual variable
        $\balpha $. The dual constraints do not have $ \Phi( \cdot) $ at all.
        A condition for applying CG is that the duality gap between the primal
        and dual problems is zero (strong duality). Generally, the primal problem must be
        convex\footnote{A Lagrange dual problem is always convex.}
        and both the primal and dual are feasible, so the Slater condition
        holds.
        In such a case, the KKT conditions
        are necessary conditions for a solution to be optimal.
        One such condition in deriving the dual \eqref{EQ:SVM_D1} from
        \eqref{EQ:SVM1} is
        \begin{equation}
            \label{EQ:KKT1}
            \bw = \textstyle \sum_{i=1}^m  y_i \alpha_i \Phi( \bx_i ).
        \end{equation}
        This KKT condition is the root of the
        {\em representer theorem} in kernel methods, which states that
        a minimizer  of a regularized empirical risk function defined over a reproducing kernel
        Hilbert space can be represented as a finite linear combination of kernel products evaluated
        on the input points.

        We can verify the optimality by checking
        the dual feasibility and KKT conditions.
        At optimality, \eqref{EQ:KKT1} must hold for all $ j $, \ie, $w_j = \sum_i y_i \alpha_i
        \wl_j( \bx_i )$ must hold for all $ j $.
        For the columns/weak learners in the current working set, \ie, $ j = 1\dots J $,
        the corresponding condition in \eqref{EQ:KKT1} is satisfied by the current solution. For the
        weak learners that are not selected yet, they do not appear in the current restricted
        optimization problem and the corresponding $ w_j = 0 $.
        It is easy to see that if
        $
            \sum_i y_i \alpha_i
            \wl_j( \bx_i ) = 0
        $
        for any $  \wl_j ( \cdot ) $ that is not in the current working set, then current solution
        is already the globally optimal one.
        So, our base learning strategy to check the optimality as well as to select the best weak
        learner $  \hat \wl(\cdot) $ is:
        \begin{equation}
            \label{EQ:WL_SV}
            {\hat \wl}(\cdot) = \argmax_{ \wl \in \calH} \;
                \Bigl\vert
                        \textstyle \sum_i y_i \alpha_i \wl(\bx_i)
                \Bigr\vert.
        \end{equation}
        Different from \eqref{EQ:WL},
        here we select the weak learner with the score
        $ \sum_i y_i \alpha_i \wl(\bx_i) $ farthest from 0, which can be negative.
        Now we show that using \eqref{EQ:WL_SV} to choose a weak learner is not heuristic
        in terms of solving the SVM problem of \eqref{EQ:SVM1}.
        \begin{result}
            At iteration $ J+1$, the weak learner selected using \eqref{EQ:WL_SV} decreases the duality gap
            the most for the current solution obtained at iteration $ J $, in terms of solving the
            SVM primal problem \eqref{EQ:SVM1} or dual \eqref{EQ:SVM_D1}.
        \end{result}
        To prove the above result, let us check the dual objective in \eqref{EQ:SVM_D1}.
        We denote the current working set (corresponding to current selected weak learners)
        by $ \calW $ and the rest by $ \bar \calW$.
        The dual objective in \eqref{EQ:SVM_D1} is
        \begin{equation}
            \label{EQ:DGap1}
            \ones^\T \balpha - \half \balpha^\T ( \bK^{ \calW } \circ \by \by ^\T ) \balpha
            - \half \balpha^\T ( \bK^{ \bar \calW } \circ \by \by ^\T ) \balpha.
        \end{equation}
        Here the $ (s, t) $ entry of  $ \bK^{\calW} $ is
        \[
        \bigl< \Phi^\calW(\bx_s ),  \Phi^{ \calW }(\bx_t )   \bigr>
            = \textstyle   \sum_{ j \in \calW}  \wl_j( \bx_s )   \wl_j( \bx_t );
        \]
        and likewise,
        $
            \bK^{\bar \calW}_{st}   =
             \sum_{ j \in \bar \calW}  \wl_j( \bx_s )   \wl_j( \bx_t ).
        $
        Clearly the sum of first terms in \eqref{EQ:DGap1} equals to the objective value of the
        primal problem with the current solution:
        $ \half \fnorm{ \bw }^2 + C \ones^\T \bxi  $.  Therefore the duality gap is the  last
        term of \eqref{EQ:DGap1}:
        $ - \half \balpha^\T ( \bK^{ \bar \calW } \circ \by \by ^\T ) \balpha
        = - \half \sum_{j \in \bar \calW}  \bigl[
                                                    \sum_i y_i \alpha_i \wl_j ( \bx_i )
                                               \bigr]^2.
        $
        Clearly minimization of this duality gap leads to  the base learning rule \eqref{EQ:WL_SV}.
        Next, we show that it can be equivalent between \eqref{EQ:WL} and \eqref{EQ:WL_SV}.
        \begin{result}
            Let us assume that
            the weak learn set $ \calH $ is negation complete; \ie,
            if $ \wl(\cdot) \in \calH $, then $ [-\wl](\cdot) \in \calH $; and vice versa.
            Here $ [-\wl](\cdot) $ means the function
            $ [-\wl](\cdot) = - (\wl(\cdot))$.
            Then to solve
            \eqref{EQ:WL_SV}, we only need to solve  \eqref{EQ:WL}.
            \label{RES:WL}
        \end{result}
        This result is straightforward. Because $ \calH $ is negation complete, if a maximizer of
        \eqref{EQ:WL_SV}, $ \hat\wl(\cdot) $,  leads to  $   \sum_iy_i \alpha_i \hat\wl(\bx_i) $
        $ < 0  $, then  $ [- \hat\wl(\cdot) ](\cdot) \in \calH $ is also a maximizer of
        \eqref{EQ:WL_SV} such that
        $   \sum_iy_i \alpha_i \hat\wl(\bx_i) $  $ > 0  $.
        Therefore we can always solve \eqref{EQ:WL} to obtain the maximizers of \eqref{EQ:WL_SV}.

        At this point, we are ready to design  CG based ensemble learning
        for solving the SVM problem, analogue to boosting, \eg, LPBoost.
        The proposed ensemble learning method,\footnote{In order not to confuse the terms, we use
        ``ensemble learning'' instead of ``boosting'' for our boosting-like algorithms.
        }
        termed \CGENS, is summarized in Algorithm \ref{ALG:1}.
        Note that at Line 6, we can use very efficient linear SVM solvers to solve either the primal
        or dual. In our experiments, we have used \liblinear \cite{liblinear}.

\def\ADot{ { $\bf \cdot$ } }%

\setcounter{AlgoLine}{0}
\linesnumbered\SetVline
\begin{algorithm}[t]
\caption{\footnotesize  \CGENS: Column generation for learning ensembles}
\centering
{\footnotesize
   \begin{minipage}[]{0.94\linewidth}
    \KwIn{
    Training data $ (\bx_i, y_i ) $, $i = 1 \cdots m$; termination
    threshold $ \epsilon > 0 $; regularization parameter $ C > 0 $;
    (optional) maximum iteration $J_{\rm max}$.
    }
   { {\bf Initialize}:
        $ J = 0 $; $ \bw = \boldsymbol 0$; $ \alpha_i = \rm const $ ($
        0 < {\rm const} < C $).
   }

   \While{ $\rm true $  }
   {
    \ADot
        Find a new weak learner $ \hat \wl(\cdot) $ by solving
        \eqref{EQ:WL};

    \ADot
        Check the termination condition:
        {\bf if} $  \sum_i y_i \alpha_i \hat \wl( \bx_i ) < \epsilon
        $, {\bf then} terminate (problem solved);

    \ADot
        Add $ \hat \wl(\cdot) $ to the restricted master problem;

    \ADot
        Solve either the primal \eqref{EQ:SVM1} or dual
        \eqref{EQ:SVM_D1}, and update $ \balpha$, $ \bw$, $b$.

    \ADot
    $ J = J + 1$; (optional) {\bf  if} $ J > J_{\rm max} $, {\bf then} terminate.
   }

\KwOut{
    The final learned model:

    \qquad \qquad $ F(\bx) = \bw^\T \Phi( \bx) $ $ + b $ $
    =  $ $ \sum_j w_j $ $ \wl_j  ( \bx) + b $.
}
\end{minipage}
}
\label{ALG:1}
\end{algorithm}

        Having shown how to solve the standard SVM problem using CG,
        we provide another example application of the proposed framework by developing a new
        multi-class ensemble method using the idea of simplex coding \cite{simplex12}.

        \section{Multi-class ensemble learning}

        As most real-world problems are inherently multi-class, multi-class learning is becoming
        increasingly important.
        Coding matrix based  boosting methods are one of the popular boosting approaches to
        multi-class classification.
        Methods in this category include AdaBoost.MO \cite{AdaMO},
        AdaBoost.OC and AdaBoost.ECC \cite{Guruswami}.
        Shen and Hao  proposed a direct approach to multi-class boosting in
        \cite{multiboost}.
        Here we proffer a new multi-class ensemble learning method based on the simplex
        least-squares SVM (SLS-SVM) introduced in \cite{simplex12}.
        SLS-SVM can be seen as a generalization of the binary LS-SVM (least-squares SVM).
        For binary classification, LS-SVM fits the decision function output to the label:
        $  \min \half \fnorm{\bw}^2  +  \tfrac{C}{2} \sum_i \xi_i^2 $
        with $  \xi_i = \bw^\T \Phi( \bx_i ) + b - y_i  $.
        In the case of multi-class classification, label $y_i \in
        \{1,2, \dots, l + 1\}$. Here we have $ l+1$ classes.
        Simplex coding  maps each class label
        to $l+1$ most separated vectors $\{
        \bc_1, \bc_2, \cdots, \bc_{l+1}
        \}$ on the unit hypersphere in
        $\mathbb{R}^{m} $.
        So we need to learn $ l $ classifiers.
        Let us assume that the simplex label coding function is $ {\spx}: \{1,\cdots, l \} $
        $ \rightarrow $ $ \{ \bc_1,\cdots,\bc_l  \} $ (see Appendix).
        The label matrix $  \bL  \in \Real^{ m \times l } $ collects training data's coded labels
          such that each row $\bL_{i:} = \spx(y_i) ^ \T$.

        SLS-SVM trains the $ l $ classifiers simultaneously by minimizing the following regularized
        problem:
        \begin{align}
            \label{EQ:SLSVM1}
            \min & \; \half \textstyle  \sum_{ \tau =1  } ^ l \fnorm{ \bw_\tau }^2 + \tfrac{C}{2}
            \sum_{i=1}^m
            \sum_{\tau=1}^l \bO_{i\tau}^2,    \notag
            \\
            \st & \;
            \bO_{i\tau} =  \bL_{i\tau} - \bw_\tau ^ \T \Phi( \bx_i ) - b_\tau.
        \end{align}
        Here the model parameters to optimize are $ \bw_\tau \in  \Real^J$, $ b_\tau \in \Real$, for
        $ \tau = 1, \cdots, l $.
        Problem \eqref{EQ:SLSVM1} can be solved by deriving its dual (see Appendix)
        and the solutions are:
        \begin{equation}\label{EQ:SLSVM_SOLUTION}
        \begin{split}
            \bb &= \Bigl(
                        \frac{\ones^\T \bS^{-1} \bL}{\ones^\T \bS^{-1} \ones}
                    \Bigr)^\T, \\
            \bU &= \bS^{-1}(\bL - \ones \bb^\T), \\
            \bw_\tau &= \H \bU_{:\tau}.
        \end{split}
        \end{equation}
        Here $\H\in\mathbb{R}^{J\times m}$ denotes the learned weak
        classifiers' response on the whole training data such that
        each column $\H_{:i}=\Phi(\bx_i)$;
        $ \bS = \H^\T \H + \tfrac{1}{C} \bI_m$;
        $ \bI_m$ is the $m\times m$ identity matrix;
        $ \bU \in \Real ^ { m \times l } $ is the dual Lagrange
        multiplier associated with the equality constraints of $\bO$.
        Note that, the inverse of $\bS$ can be computed efficiently incrementally
        (see the supplementary document).

        Here one of the KKT conditions that CG relies on is (last equation of
        \eqref{EQ:SLSVM_SOLUTION}):
        \begin{equation}
            \label{EQ:KKT2}
            \bw_\tau = \textstyle  \sum_i  \bU_{i\tau}   \Phi(  \bx_i ),
        \end{equation}

        Similar to the binary case,
        the subproblem for generating weak learners is
        \begin{equation}
            \label{EQ:WL2}
               {\hat \wl}(\cdot) = \argmax_{ \tau=1\cdots l,
                                             \wl \in \calH} \;
                \textstyle \sum_i  \bU_{i\tau} \wl(\bx_i).
        \end{equation}
        For the same reason as in Result \ref{RES:WL}, we have removed the absolute operation
        without changing the essential problem.
        A subtle difference from the binary classification
        is that we pick the best weak learner across all the $ l $ classifiers.

\def\ADot{ { $\bf \cdot$ } }%

\setcounter{AlgoLine}{0}
\linesnumbered\SetVline
\begin{algorithm}[t]
\caption{\footnotesize \CGENS-SLS: Simplex coding multi-class
ensembles}
\centering
{\footnotesize
   \begin{minipage}[]{0.94\linewidth}
    \KwIn{
    Training data $ (\bx_i, y_i ) $, $i = 1 \cdots m$; termination
    threshold $ \epsilon > 0 $; regularization parameter $ C > 0 $;
    (optional) maximum iteration $J_{\rm max}$.
    }
   { {\bf Initialize}:
        $ J = 0 $;  $\bL_{i:}=\spx(y_i)^\T$;
        Assign a positive constant to each element of $ \bU_{i\tau}$.
        $ \bw_\tau = \boldsymbol 0$, $b_\tau=0$, $\tau = 1 \cdots l$.
   }

   \While{ $\rm true $  }
   {
    \ADot
        Find a new weak learner $ \hat \wl(\cdot) $ by solving
        \eqref{EQ:WL2};

    \ADot
        Check the termination condition:
        {\bf if} \, $\mathop{\max}_\tau   \sum_i \hat \wl(\bx_i)U_{i\tau}
                           < \epsilon $, {\bf then} terminate (problem solved);

    \ADot
        Add $ \hat \wl(\cdot) $ to the restricted master problem;

    \ADot
         Update $\H$, $ \bU$, $\bw_\tau$
        using \eqref{EQ:SLSVM_SOLUTION} (solving \eqref{EQ:SLSVM1});

    \ADot
    $ J = J + 1$; (optional) {\bf  if} $ J > J_{\rm max} $, {\bf then} terminate.
   }

\KwOut{
    The learned $ l $ classifiers:
    $ F_\tau( \bx ) = \bw_\tau ^\T \Phi(\bx) + b_\tau  $, $ \tau =
    1,\cdots,l$;
    with $ \Phi(\bx) = [ \wl_1(\bx), \cdots, \wl_J(\bx) ]^\T $.
}
\end{minipage}
}
\label{ALG:LSSVM}
\end{algorithm}

We summarize our multi-class ensemble method in Algorithm \ref{ALG:LSSVM}.
The output is the $ l $ classifiers:
\begin{equation}
\begin{split}
    F(\bx) &= \left[ \bw_1^\T \Phi(\bx)+b_1,\cdots, \bw_l^\T
    \Phi(\bx)+b_l \right],
\end{split}
\end{equation}
The classification rule assigns the label
\begin{equation}
      \argmax_{y=1, \cdots, l+1} \langle F(\bx),\bc_y \rangle
\end{equation}
to the test datum $ \bx $.

        \section{Experiments}
        We run experiments on binary and multi-class classification problems, and compare our methods against classical boosting methods.

\subsection{Binary classification}
        We conduct experiments on synthetic as well as real datasets, namely, 2D toy, 13 UCI benchmark datasets\footnote{\url{http://www.raetschlab.org/ Members/ raetsch/ benchmark}},
        and then on several vision
        tasks such as digits recognition, pedestrian detection \etc.

\begin{figure}
    \centering
        \includegraphics[width=0.4\textwidth]{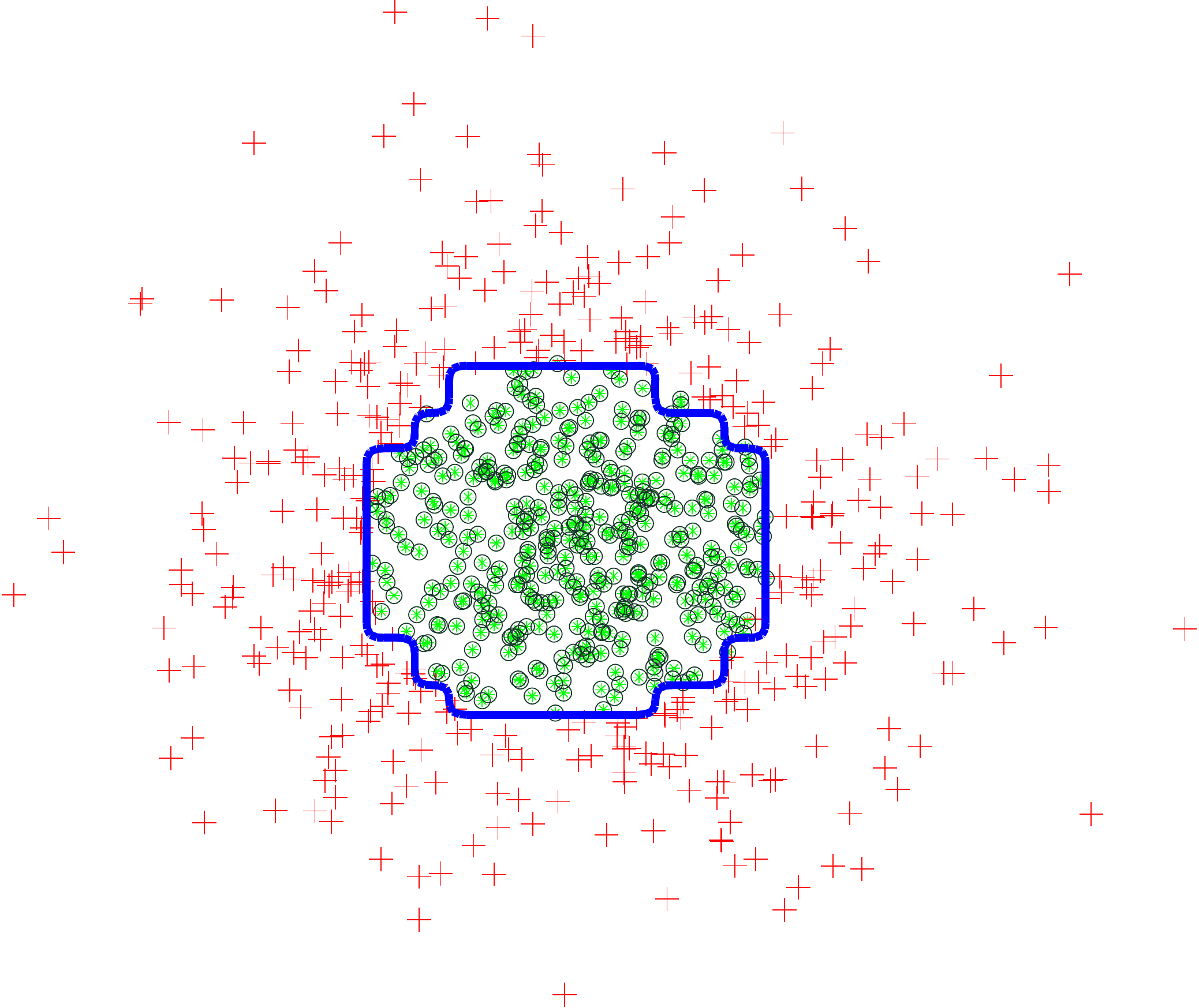}
        \includegraphics[width=0.48\textwidth]{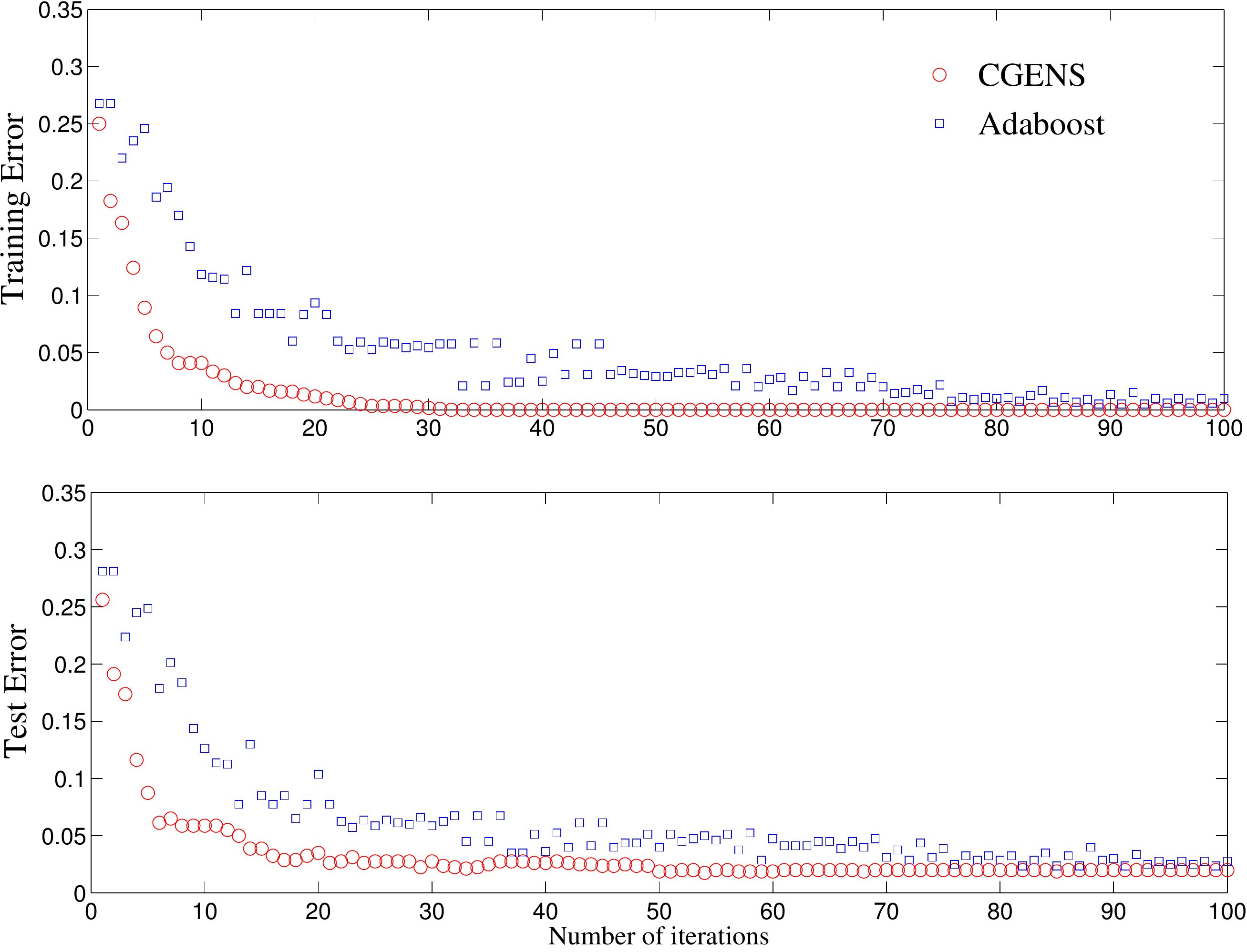}
    \caption {(left) Decision boundary of the proposed method on toy data;
    (right) training and testing errors of AdaBoost and \CGENS
    versus numbers of iterations. \CGENS converges faster than AdaBoost.
    }
      \label{figToy2000}
\end{figure}

       \textbf{2D toy data}
         The data are generated by sampling points from a 2D Gaussian distribution.
        All points within a certain radius belong to one class and all outside belong to the other class.
        The decision boundary obtained by \CGENS with decision stumps is plotted
        in Fig. \ref{figToy2000}.
        As can be seen, our method converges  faster than
        AdaBoost because \CGENS is fully corrective.

\textbf{UCI benchmark}
        For the UCI experiment, we use
        three different weak learners, namely, decision stumps, perceptrons and Fourier weak
        learners, with each
        compared with the corresponding kernel SVMs and other boosting methods.

        We use decision stumps (stump kernel for SVM) in this experiment.
        We compared our method with AdaBoost, LPBoost, all using decision stumps as the weak learner.
        All the parameters are chosen using
        $5$-fold cross validation. The maximum iteration for AdaBoost, LPBoost \cite{LPB}
        and our \CGENS are searched from $\{25, 50, 100, 250, 500\}$.
        Results of SVMs with the stump kernel are also reported \cite{Lin2008}. Results are the
        average of 5 random splits on each dataset.  From Table \ref{tabAccUCIstump}, we can see
        that overall, all the methods achieve comparable accuracy, with \CGENS being marginally the
        best, and SVM the second best.

\begin{table}
\centering
\scalebox{0.96}{
    \begin{tabular}{r|c|c|c|c}
        \hline
                & AdaBoost & LPBoost & SVM & {\CGENS} \\
        \hline\hline
        banana & 28.38$\pm$0.55  & 27.28$\pm$0.25  & 27.16$\pm$1.70  & \textbf{26.97$\pm$1.35} \\
        b-cancer & \textbf{28.83$\pm$6.12}  &29.90$\pm$1.84  &29.61$\pm$1.69  & 30.13$\pm$2.13  \\
        diabetes & 25.13$\pm$1.48  & 25.66$\pm$2.41  &25.53$\pm$1.32  & \textbf{25.07$\pm$2.35}  \\
        f-solar & 32.85$\pm$1.44  & 31.98$\pm$1.69  & 32.30$\pm$2.45  & \textbf{31.80$\pm$1.70}  \\
        german & 26.27$\pm$2.14  & \textbf{22.88$\pm$1.74}  & \textbf{22.87$\pm$1.74}  & 22.87$\pm$2.39  \\
        heart & 18.80$\pm$4.60  & 18.60$\pm$4.45  & \textbf{17.00$\pm$3.39}  & 18.40$\pm$4.51  \\
        image & 2.55$\pm$0.67  & \textbf{2.44$\pm$0.63}  & 3.62$\pm$0.58  & 2.85$\pm$0.45 \\
        ringnorm & 8.72$\pm$0.66  & 8.98$\pm$0.88  & \textbf{3.93$\pm$1.66}  & 5.71$\pm$0.42 \\
        splice & 8.48$\pm$0.57  & 8.30$\pm$0.87  & 7.56$\pm$0.24  & \textbf{6.23$\pm$0.47} \\
        thyroid & 5.87$\pm$4.58  & 6.94$\pm$5.56  & 5.87$\pm$3.60  & \textbf{5.60$\pm$3.45} \\
        titanic & 24.53$\pm$2.86  & 22.38$\pm$0.33  & \textbf{22.30$\pm$0.53}  & \textbf{22.30$\pm$0.53} \\
        twonorm & 4.66$\pm$0.34  & 5.32$\pm$0.73  & \textbf{3.07$\pm$0.54}  & 4.40$\pm$0.46 \\
        waveform & 12.86$\pm$0.46  & 12.96$\pm$0.59  & 12.84$\pm$2.32  & \textbf{12.63$\pm$0.67} \\
        \hline
    \end{tabular}
    }
    \vspace{.3cm}
    \caption{Mean and standard deviation of test errors ($\%$) on 13 UCI datasets
    using stumps (stump kernel for SVM).}
    \label{tabAccUCIstump}
\end{table}

      In the second experiment, we compare our method (using 500 weak learners) against several
      other methods such as SVM using (1) perceptrons $\sign(
      \theta^\T \bx - \kappa  )$ as weak learners and the perceptron kernel for SVM, and (2) Fourier
      cosine functions
      \cite{Rahimi07}
      $ \cos( \theta^\T \bx - \kappa  ) $ as weak learners and Gaussian RBF kernel for SVM.
      We did not optimize the weak learner's parameters $\{ \theta, \kappa \} $. Instead, we sample
      2000 pairs of $\{ \theta, \kappa \} $ according to their distributions as described in
      \cite{Lin2008} and \cite{Rahimi07}, and then pick the one that maximizes the weak learner
      selection criterion in Equ.\ \eqref{EQ:WL_SV}.

      In the case of the perceptron kernel, same as the decision stump kernel, it is parameter free.
      In the case of Gaussian RBF kernel (corresponding Fourier  weak learners),
      there is Gaussian bandwidth parameter $ \sigma $.
      Here $ \theta $ in  Fourier is sampled from a Gaussian distribution with the same bandwidth.
      We cross validate this bandwidth parameter $ \sigma $ with the SVM and use the same $ \sigma $
      for sampling Fourier  weak learners for use in \CGENS. Ideally one can cross validate
      $ \sigma $ with \CGENS, which needs extra computation overhead.
       This might be the reason  why RBF SVM is slightly  better \CGENS
       with Fourier weak learners as shown  in Table \ref{tabAccUCIrbf},
       because \CGENS uses the optimal $ \sigma $ of SVM.
       While in the case of perceptrons, \CGENS performs on par with SVM. Note that as expected, in
       general, \CGENS and SVM again outperform AdaBoost and LPBoost.

       We have also compared our \CGENS with RFF \cite{Rahimi07}. Although \CGENS uses 500
       features (weak learners)
       and  RFF uses 2000 features, \CGENS still slightly outperforms RFF.

\begin{table*}

\centering
\scalebox{0.9}{
    \begin{tabular}{ r | c | c | c | c | c | c  | c }
    \hline
        \multirow{2}{*} {{   }} & \multicolumn{3}{ c| } {{Fourier weak
        learner$/$RBF kernel}} &
        \multicolumn{4}{ c }{{Perceptron$/$Perceptron kernel} } \\
        \cline{2-8}
        & RFF \cite{Rahimi07}   &{SVM} & \CGENS &     AdaBoost  & {LPBoost} & {SVM}  & {\CGENS} \\
        \hline
        \hline
        banana & \textbf{11.41$\pm$0.84} & 11.64$\pm$0.66  & 11.74$\pm$0.67  &      11.40$\pm$0.48  & 12.52$\pm$1.15  & 11.72$\pm$0.78  & \textbf{10.88$\pm$0.28}  \\
        b-cancer & 28.83$\pm$4.35 & 28.31$\pm$4.63 & \textbf{27.79$\pm$3.26}  &    30.91$\pm$5.91  & 30.68$\pm$4.27  & 29.87$\pm$6.30  & \textbf{28.83$\pm$4.81}  \\
        diabetes & 24.27$\pm$0.68 & 24.33$\pm$1.51 & \textbf{23.87$\pm$2.34}  &    26.87$\pm$3.01  & 25.54$\pm$1.58  & \textbf{24.07$\pm$1.40}  & 24.73$\pm$1.55  \\
        f-solar & \textbf{31.85$\pm$1.66} & \textbf{31.85$\pm$1.66} & 32.15$\pm$1.45  &  34.80$\pm$1.57  & 33.96$\pm$1.50  & 33.10$\pm$2.02  & \textbf{32.05$\pm$1.59}  \\
        german & 23.53$\pm$2.58 & \textbf{22.80$\pm$2.06} & 24.13$\pm$2.35  &     26.00$\pm$1.05  & 24.04$\pm$2.59  & 24.33$\pm$3.79  & \textbf{21.60$\pm$2.70}  \\
        heart & \textbf{17.00$\pm$4.53} & 17.60$\pm$4.67 & 17.80$\pm$4.82  &    20.60$\pm$6.43  & 20.06$\pm$4.83  & \textbf{17.80$\pm$2.95}  & 18.60$\pm$6.07  \\
        image & 3.80$\pm$1.68 & \textbf{2.99$\pm$0.49}  & 3.90$\pm$0.98  &     2.75$\pm$0.50  & 3.06$\pm$0.51  & 3.19$\pm$0.44  & \textbf{2.50$\pm$0.52}  \\
        ringnorm & 2.03$\pm$0.13 & \textbf{1.66$\pm$0.08}  & 2.47$\pm$0.21  &     4.99$\pm$0.50  & 7.96$\pm$0.45  & \textbf{2.15$\pm$0.23}  & 3.02$\pm$0.21  \\
        splice & 13.14$\pm$0.89 & \textbf{11.32$\pm$0.77} & 12.02$\pm$0.40  &     13.87$\pm$0.84  & 15.28$\pm$0.84  & \textbf{12.46$\pm$1.28}  & 13.51$\pm$1.08  \\
        thyroid & 4.27$\pm$2.19 & \textbf{3.73$\pm$1.46} & 3.73$\pm$2.56  &    \textbf{3.47$\pm$1.19}  & 7.46$\pm$4.59  & 6.93$\pm$3.04  & 4.80$\pm$3.21  \\
        titanic & 22.42$\pm$0.66 & 22.42$\pm$0.66 & \textbf{22.37$\pm$0.73}  &     24.05$\pm$1.89  & 22.56$\pm$1.89  & 22.56$\pm$0.75  & \textbf{22.09$\pm$0.83}  \\
        twonorm & 3.71$\pm$1.59 & \textbf{2.81$\pm$0.27} & 3.00$\pm$0.61  &     3.27$\pm$0.36  & 3.42$\pm$0.48  & \textbf{2.45$\pm$0.13}  & 3.19$\pm$0.49  \\
        waveform & 10.80$\pm$0.37 & 10.96$\pm$0.32  & \textbf{10.47$\pm$0.23}  &    11.20$\pm$0.23  & 11.98$\pm$0.55  & 11.96$\pm$0.63  & \textbf{10.80$\pm$0.33}  \\
    \hline
    \end{tabular}
    }
    \vspace{.3cm}
    \caption{Test errors ($\%$) on the 13 UCI benchmarks using
    Fourier weak learns (corresponding to Gaussian RBF kernels for SVM)
    and perceptrons weak learners.}
    \label{tabAccUCIrbf}
\end{table*}

    {\bf Computation efficiency of \CGENS}
    We evaluate the computation efficiency of the proposed \CGENS.
    As mentioned, At each CG iteration of \CGENS, we need to solve a linear SVM and therefore we
    can take advantage of off-the-shelf fast linear SVM solvers. Here we have used \liblinear.
    We compare \CGENS against LPBoost because both are fully-corrective boosting.
    At each iteration, LPBoost needs to solve a linear program.
    We use the state-of-the-art commercial solver Mosek \cite{Mosek}  to solve the  dual problem \eqref{EQ:LPB_Dual}.
    The dual
    problem has less number of variables than the primal \eqref{EQ:LPB1}. Thus it more efficient for
    Mosek to solve the dual problem.

\begin{figure}[t]
    \centering
        \includegraphics[width=0.4\textwidth]{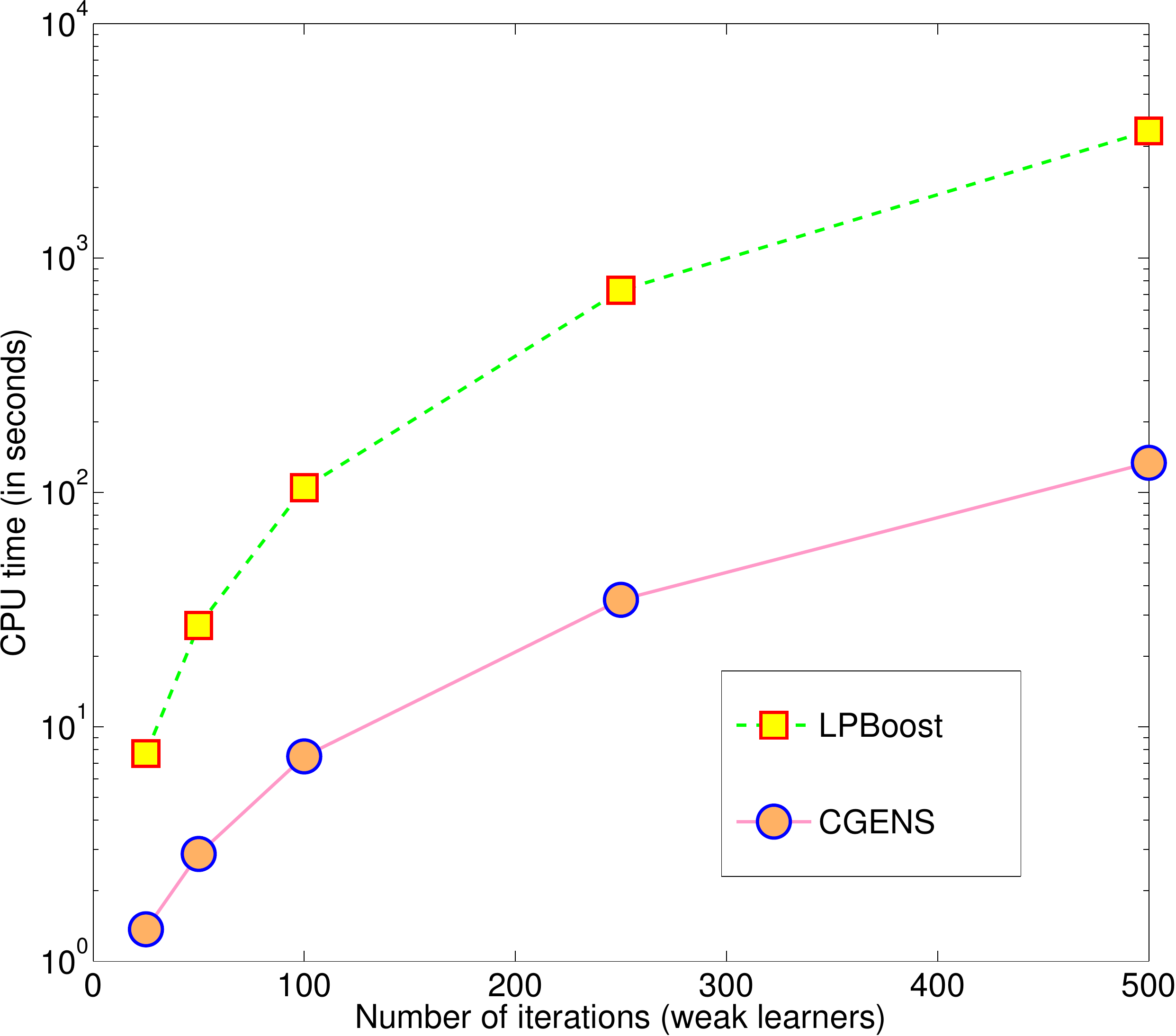}
        \includegraphics[width=0.4\textwidth]{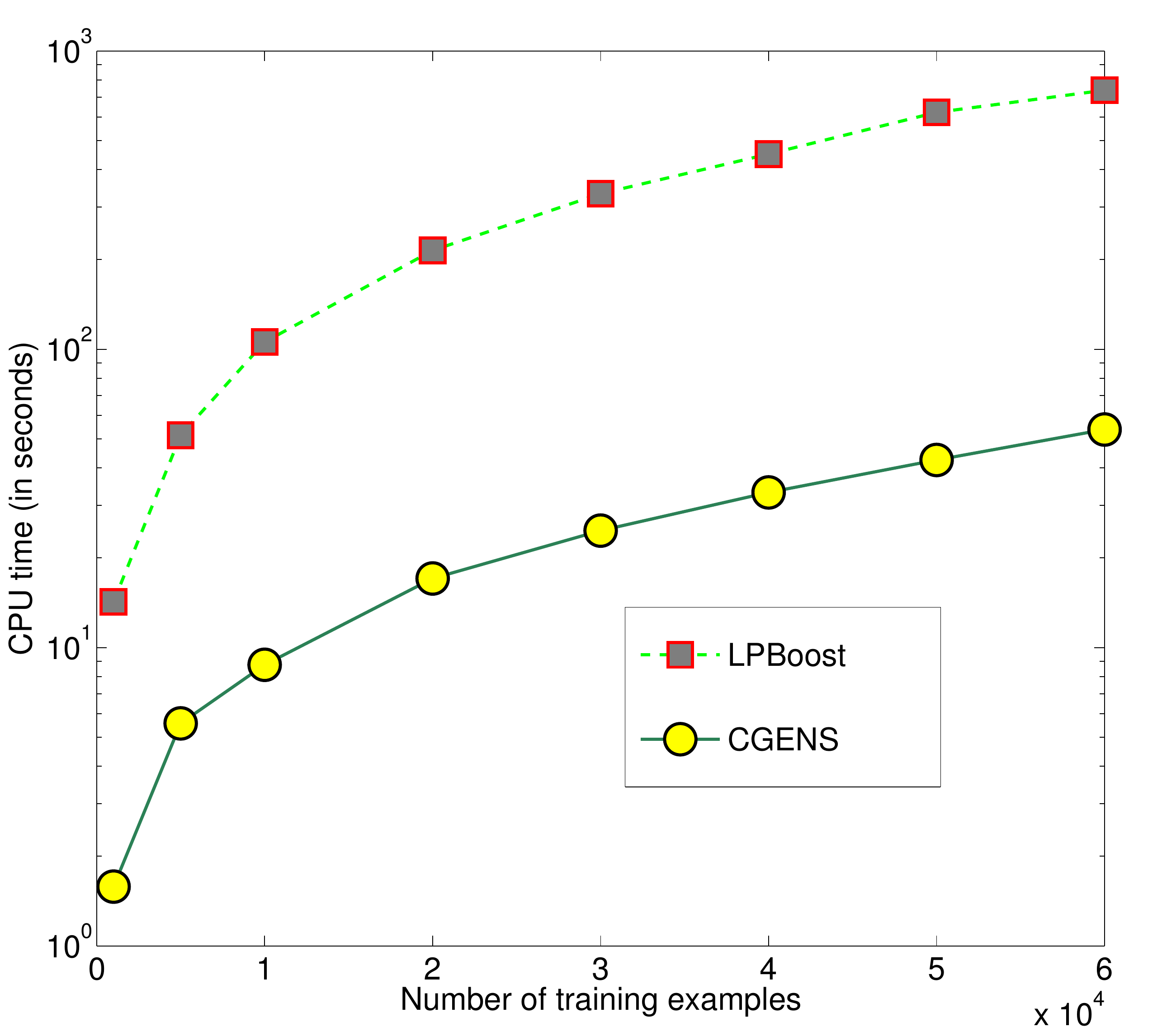}
        \vspace{.3cm}
        \caption{Training time of LPBoost \cite{LPB} and our \CGENS in log-scale.
        (top) CPU time vs.\ varying number of iterations. At iteration 500, \CGENS is about 15 times
        faster than LPBoost;
        and (bottom) CPU time vs.\ varying number of training examples. When 60,000 training data
        are used, \CGENS is about 26 times faster.
    }
      \label{fig:speed1}
\end{figure}

    We run experiments on a standard desktop using  the MNIST data to differentiate odd from even digits.
    First we vary the number of iterations (selected weak learners) while fixing the number of
    training data to be $10^4$. For the second one, we vary the number of training data and fix the
    iteration number to be 100.
    Fig.\ \ref{fig:speed1} reports the comparison results. Note that the CPU time includes the
    training time of decision stumps. Overall, \CGENS is orders of magnitude faster than LPBoost.

\subsection{Multi-class classification}
To demonstrate the potential effectiveness of the proposed ensemble learning
method in multi-class classification task, we test the proposed \lssvm
algorithm both on UCI and image benchmark datasets.
For fair comparison, we
focus on the multi-class algorithms using binary weak learners,
including  AdaBoost.ECC
\cite{Guruswami} , AdaBoost.MH \cite{AdaMO} and MultiBoost \cite{multiboost} using the exponential
loss.

The proposed \lssvm method is more related to MultiBoost in the sense
that both use the fully-corrective boosting framework, yet it employs an LS
regression-type
formulation of multi-class classifier and a closed-form solution exists for the sub optimization problem during each
iteration.  For all boosting algorithms, decision stumps are chose as
the weak learners due to its simplicity and the controlled
complexity.  Similar to the binary classification experiments, the
maximum number of iteration is set to 500. The regularization
parameters in our \lssvm and MultiBoost \cite{multiboost} are both determined by 5-fold cross validation.
We first test the proposed \lssvm on 7 UCI datasets, and then on several vision tasks. The results are summarized in Table \ref{table:lssvm}.

\textbf{UCI datasets}
For each of the 7 dataset, all
samples are randomly divided into 75\% for training and 25\% for test,
regardless of the existence of a pre-specified split.
Each algorithm is evaluated 10 times and the average results are reported.

\begin{table}
\centering
\scalebox{.9}{
\begin{tabular}{r|c|c|c|c}
\hline
& AdaBoost.ECC & AdaBoost.MH & MultiBoost \cite{multiboost} & \lssvm \\
\hline
\hline
wine$^*$  &3.4$\pm$2.9  &3.6$\pm$2.5 &3.0$\pm$2.9 &\textbf{2.3$\pm$1.9} \\
iris$^*$   &7.3$\pm$2.1 &6.2$\pm$1.7  &\textbf{5.7$\pm$2.2} &6.0$\pm$4.0 \\
glass$^*$  &24.2$\pm$5.3 &\textbf{23.2$\pm$4.7} &31.5$\pm$8.6 &25.9$\pm$5.8\\
vehicle$^*$  &22.3$\pm$2.1 &\textbf{22.0$\pm$ 1.8} &30.3$\pm$3.2 &22.2$\pm$1.5\\
DNA$^*$  &6.4$\pm$0.5 &5.9$\pm$0.5  &6.1$\pm$0.4 &\textbf{5.4$\pm$1.2} \\
vowel$^*$   &31.1$\pm$3.2 &20.1$\pm$ 2.6  &32.7$\pm$11.4 &\textbf{19.3$\pm$1.1} \\
segment$^*$ &4.2$\pm$0.8 &\textbf{2.3$\pm$ 0.5}  &3.6$\pm$0.6 &2.6$\pm$0.5 \\
MNIST &12.1$\pm$0.3 &10.9$\pm$0.2 &\textbf{10.4$\pm$0.4} &12.0$\pm$0.2 \\
USPS &6.1$\pm$0.7 &\textbf{5.6$\pm$0.3}  &8.5$\pm$0.4 &6.3$\pm$0.5 \\
PASCAL07  &56.5$\pm$0.2 &54.5$\pm$0.8  &55.9$\pm$1.4 &\textbf{53.6$\pm$0.8} \\
LabelMe &27.3$\pm$0.4 &25.4$\pm$0.4  &\textbf{25.0$\pm$0.4} &26.2$\pm$0.1 \\
CIFAR10 &52.0$\pm$0.6 &49.5$\pm$0.6  &50.8$\pm$0.5 &\textbf{47.1$\pm$0.4} \\
Scene15 &26.5$\pm$0.9 &24.9$\pm$0.6  &27.8$\pm$0.7 &\textbf{23.9$\pm$0.3}\\
SUN &60.8$\pm$1.1  &56.4$\pm$1.3 &61.1$\pm$0.8 &\textbf{53.9$\pm$1.5} \\
\hline
\end{tabular}
}
\vspace{.3cm}
\caption{Comparison of test errors (\%) on 7 UCI and 7 vision datasets. The results of the UCI datasets (marked with $*$) are averaged over 10 different runs while all the others are averaged over 5 tests.
Decision stumps are used as weak learners here.
The best results are bold faced.}
\label{table:lssvm}
\end{table}

\textbf{Handwritten digit recognition}
Three handwritten digit datasets are evaluated here, namely, MNIST, USPS and PENDIGITS. For MNIST, we randomly sample 1000 examples from each class for training and use the original test set of 10000 examples for test.
For USPS, %
we randomly select $75\%$ for training and the rest for test.

\textbf{Image classification}
We then apply the proposed \lssvm for image classification on several datasets: PASCAL07 \cite{pascal07}, LabelMe \cite{labelme} and CIFAR10.
For PASCAL07, we use 5 types of features provided in
\cite{multimodal}. For LabelMe, we use the LabelMe-12-50k\footnote{http://www.ais.uni-bonn.de/download/datasets.html} subset and
generate the GIST \cite{gist} features. Images which have more than
one class labels are excluded for these two datasets. We use 70\% of
the data for training, and the rest for test. As for CIFAR10\footnote{http://www.cs.toronto.edu/~kriz/cifar.html}, we also use the GIST \cite{gist} features and use the provided test and training sets.

\textbf{Scene recognition}
The Scene15 dataset consists of 4,485 images of 9 outdoor scenes and 6 indoor
scenes. We randomly select 100 images per class for training and the
rest for test. Each image is divided into 31 sub-windows, each of
which is represented as a histogram of 200 visual code words,
leading to a 6200D representation.
For the SUN dataset, we construct a subset of the original dataset
containing 25 categories, where the top 200 images are selected from
each category. For the subset, we randomly select 80\% data for
training and the rest for test. HOG features described in \cite{sun}
are used as the image feature.

\begin{figure*}
    \centering
    \includegraphics[width=0.186\textwidth, height=0.15\textwidth]{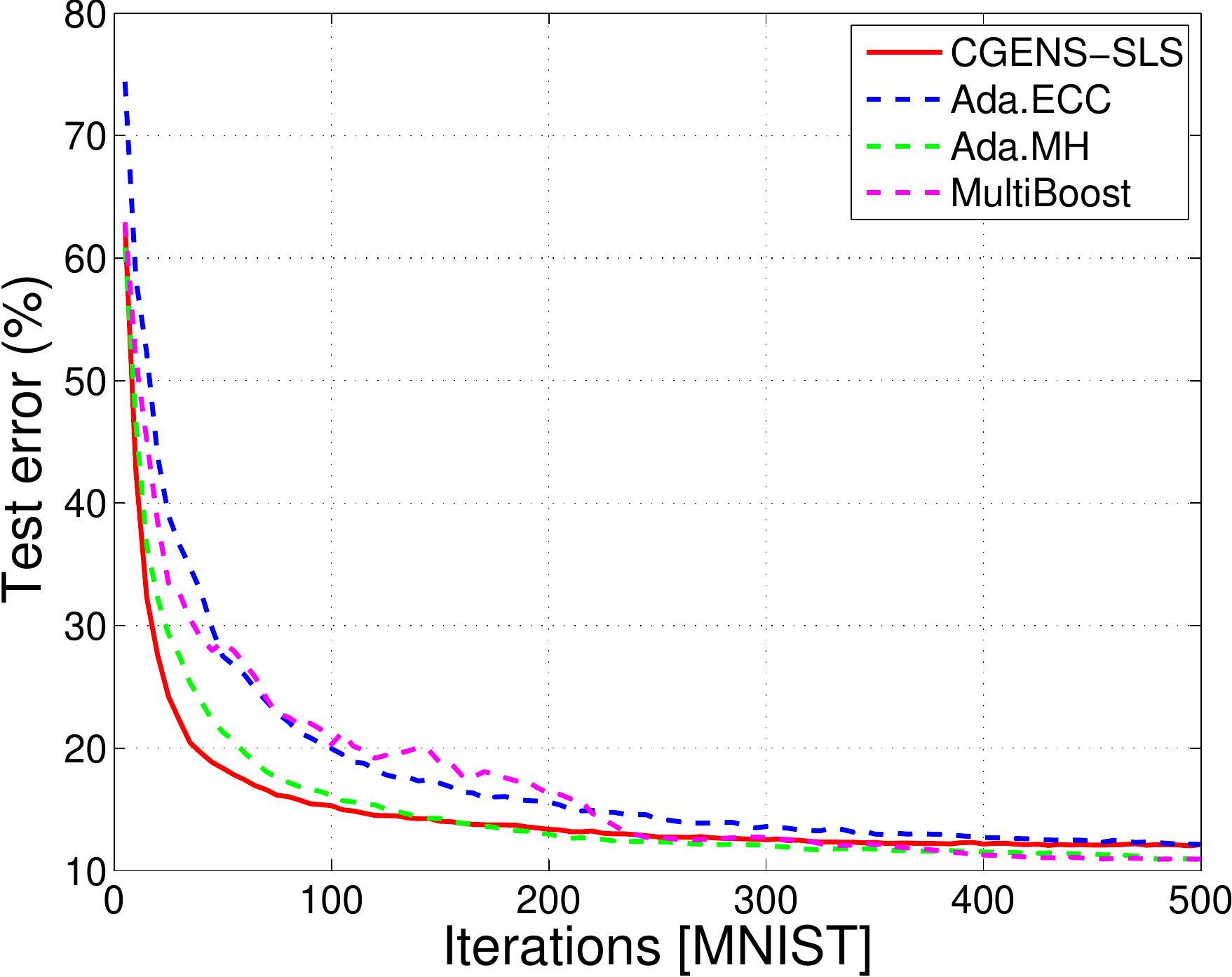}
    \includegraphics[width=0.186\textwidth, height=0.15\textwidth]{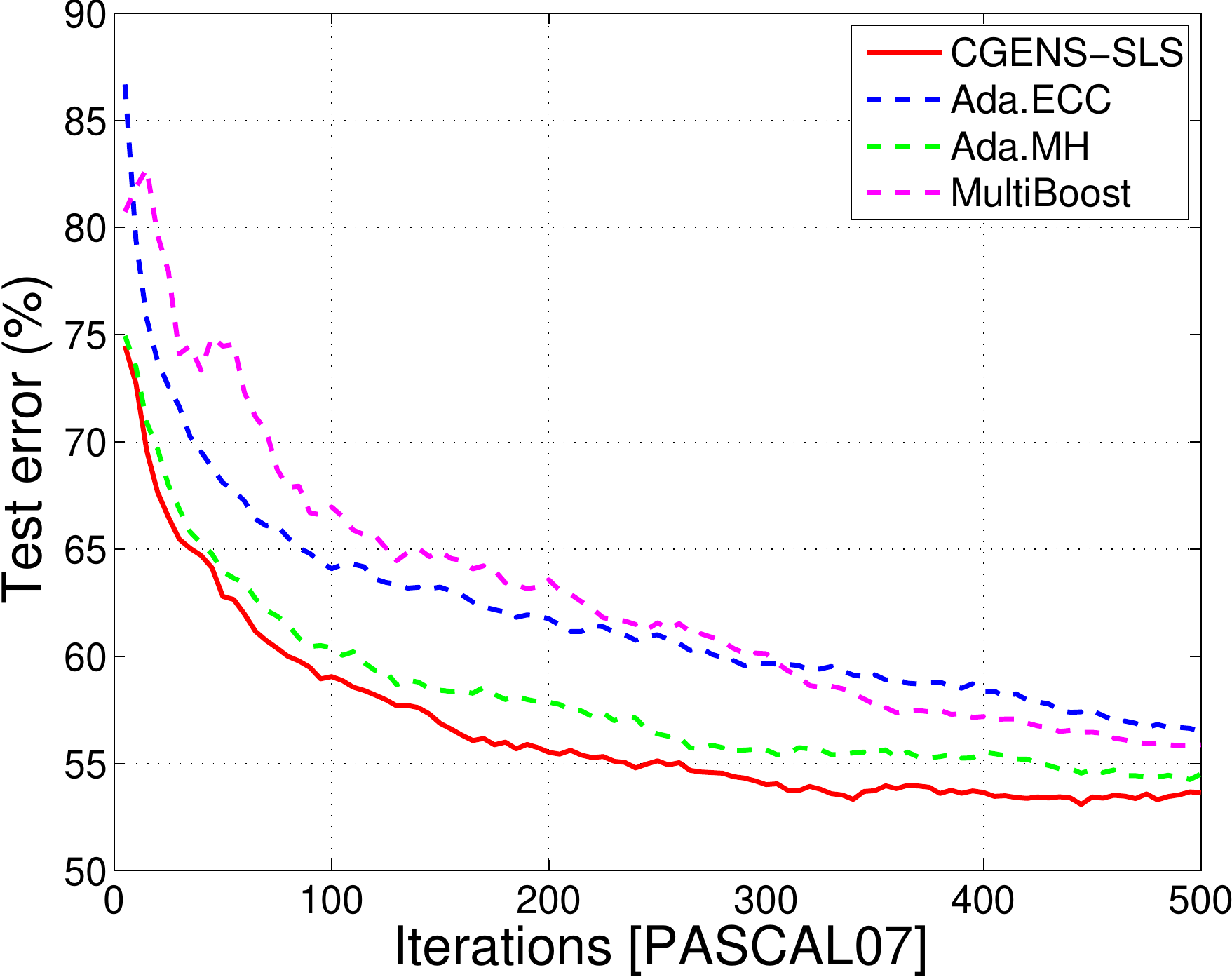}
    \includegraphics[width=0.186\textwidth, height=0.15\textwidth]{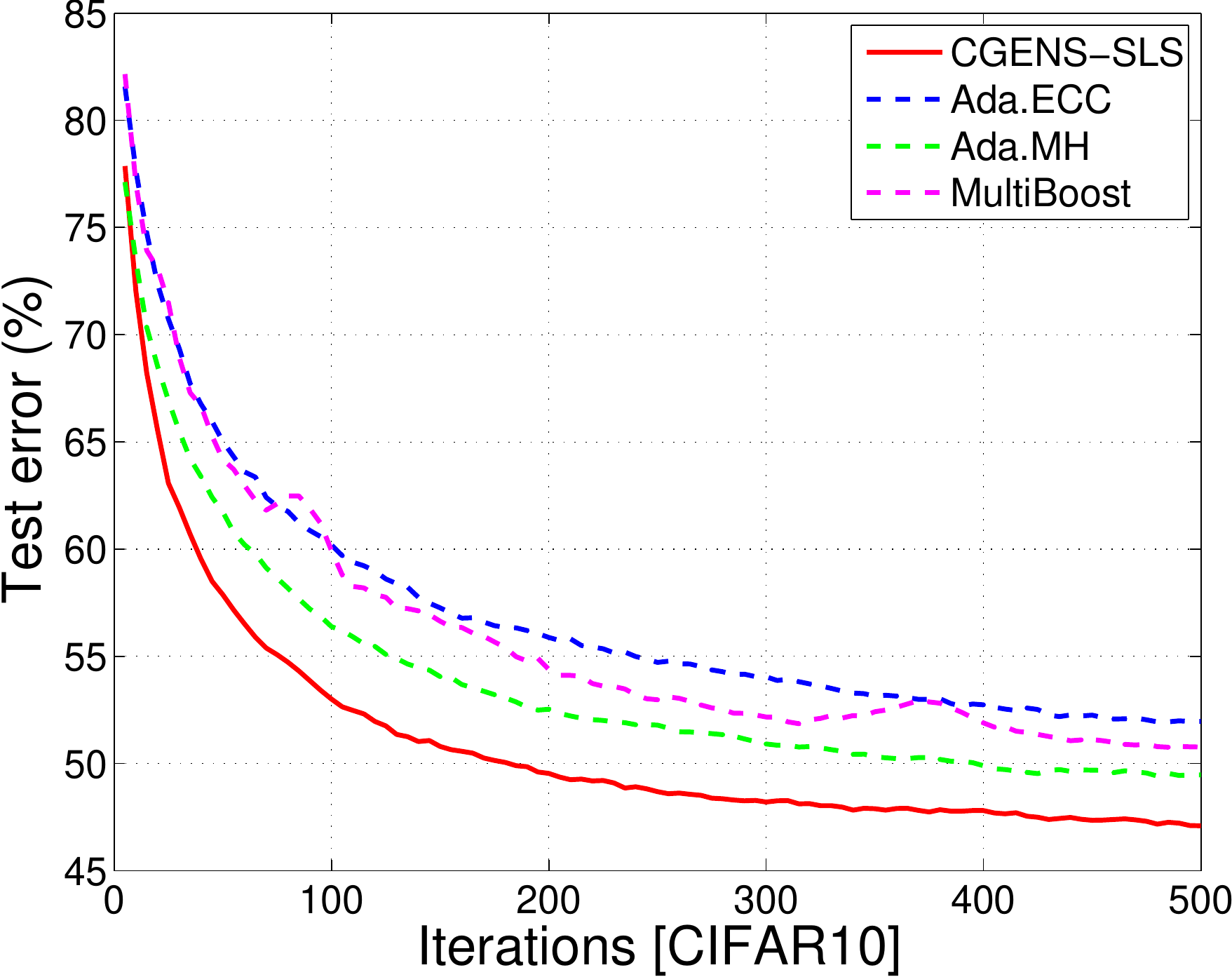}
    \includegraphics[width=0.186\textwidth, height=0.15\textwidth]{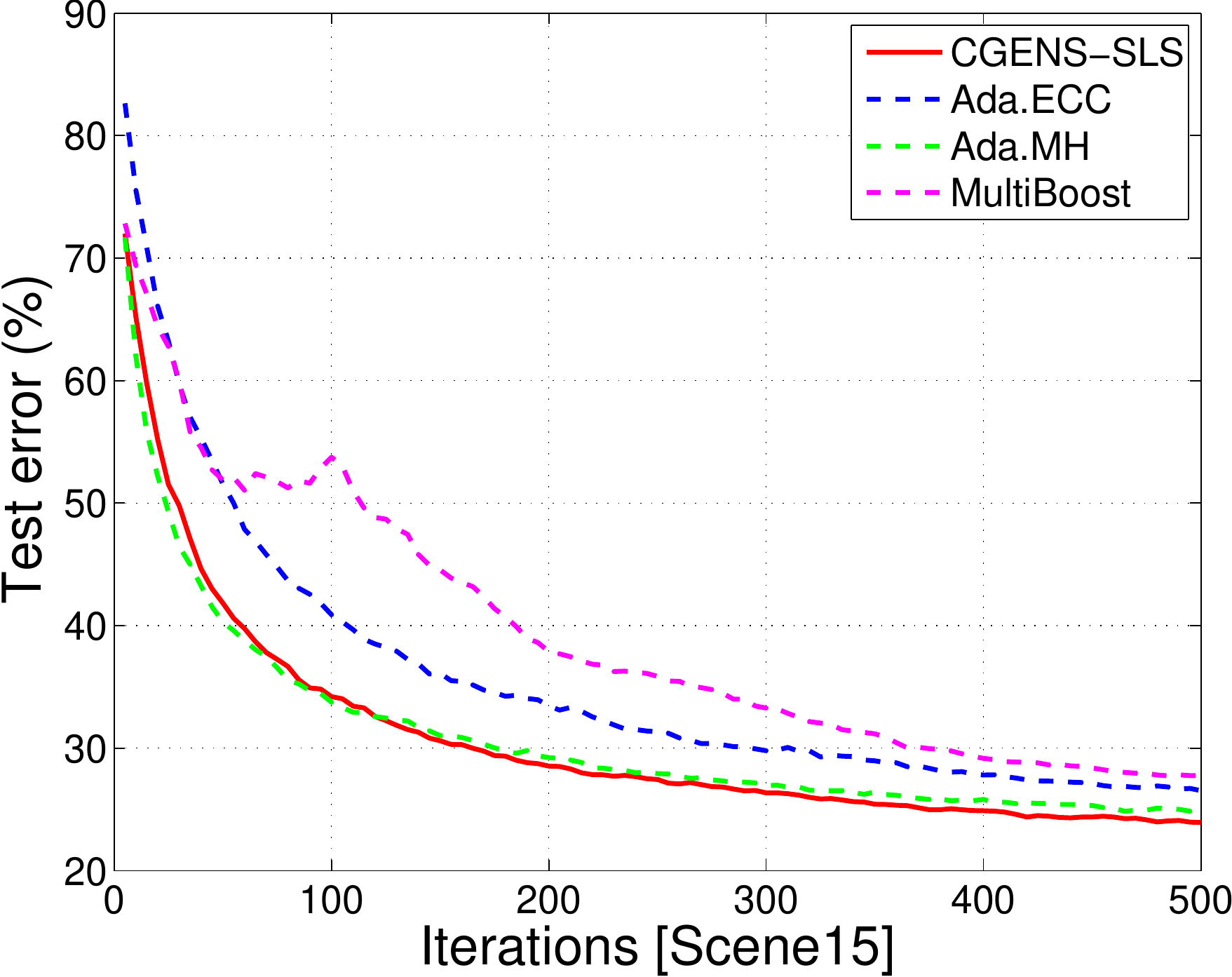}
    \includegraphics[width=0.186\textwidth, height=0.15\textwidth]{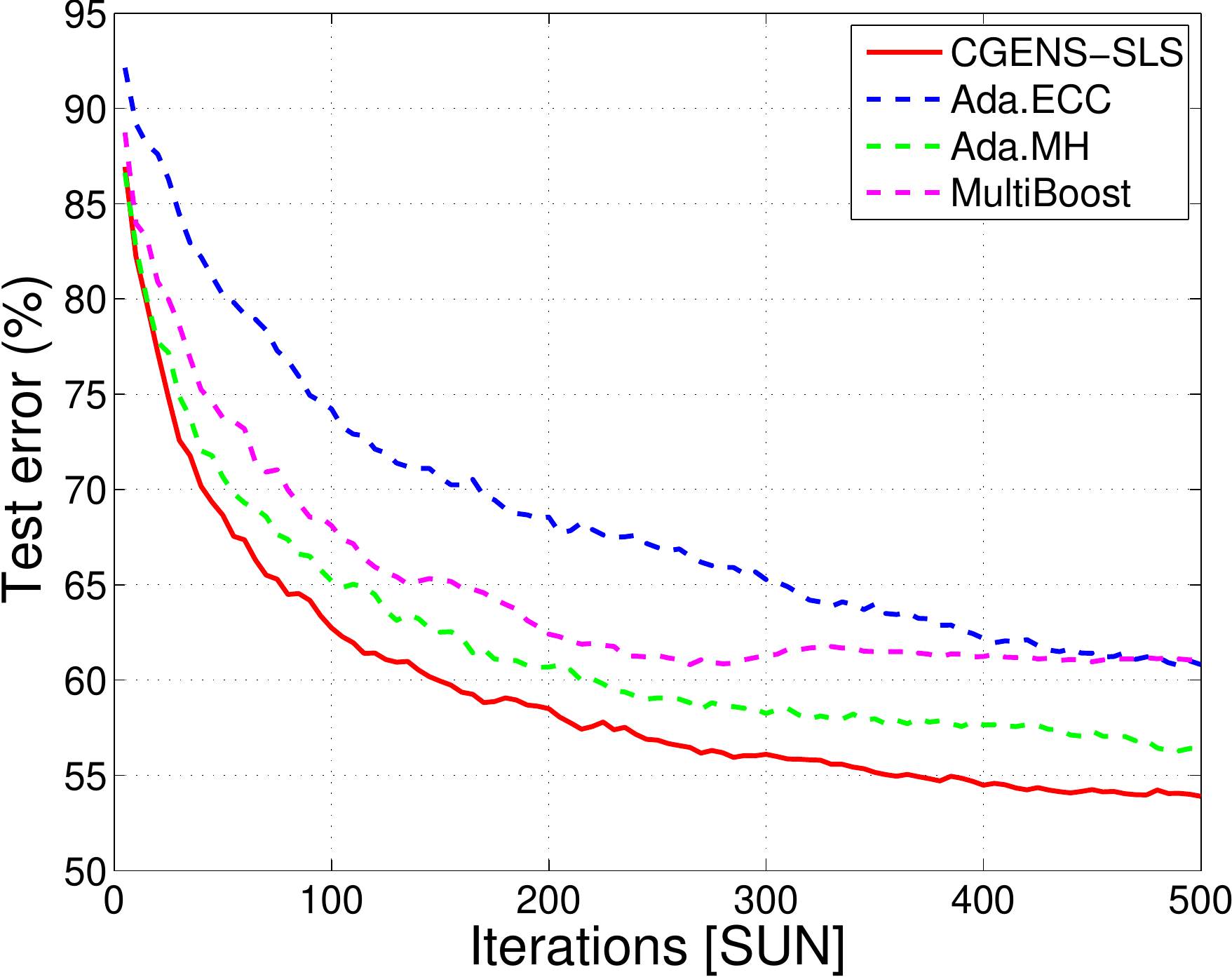} \\
    \vspace{.2cm}
    \includegraphics[width=0.186\textwidth, height=0.15\textwidth]{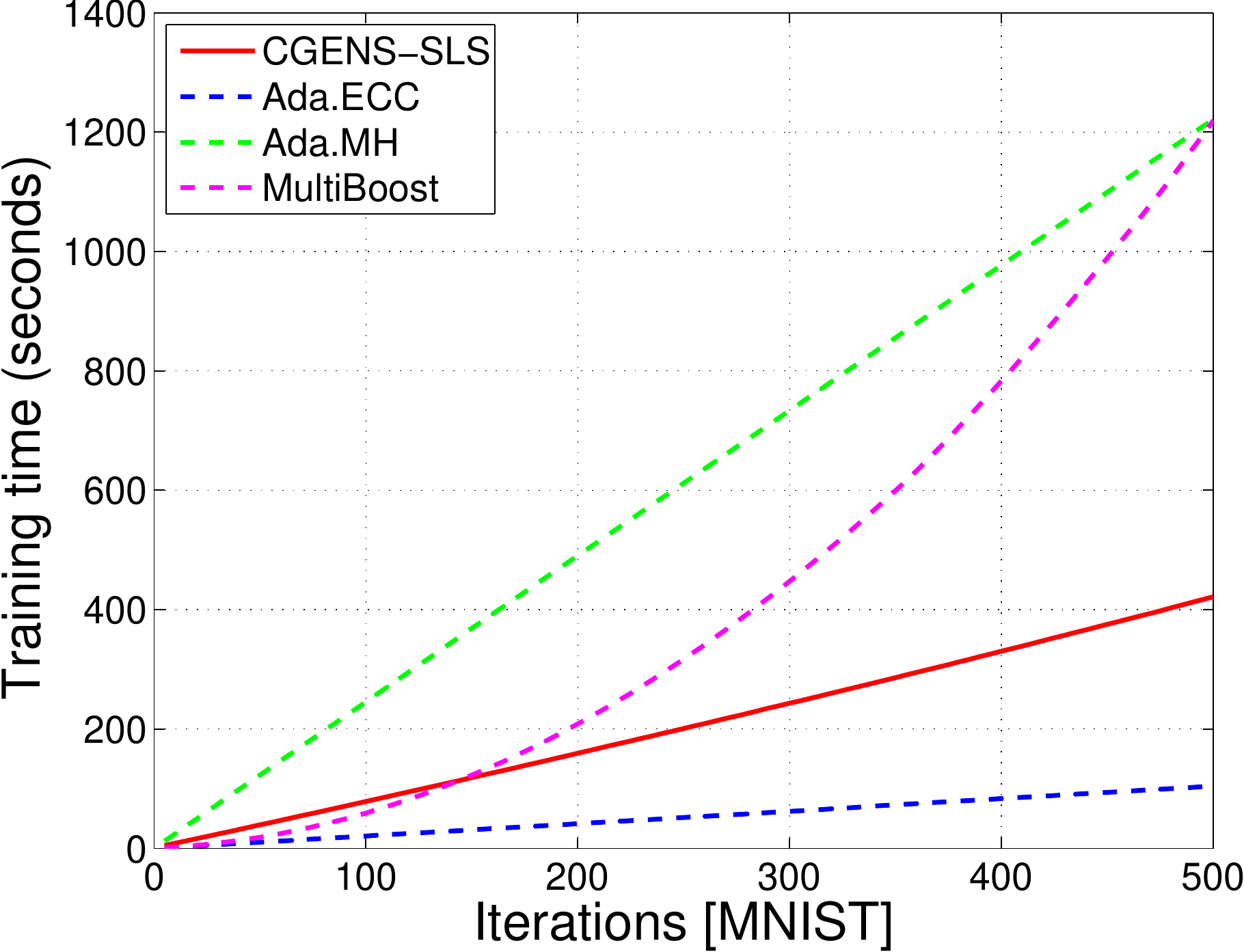}
    \includegraphics[width=0.186\textwidth, height=0.15\textwidth]{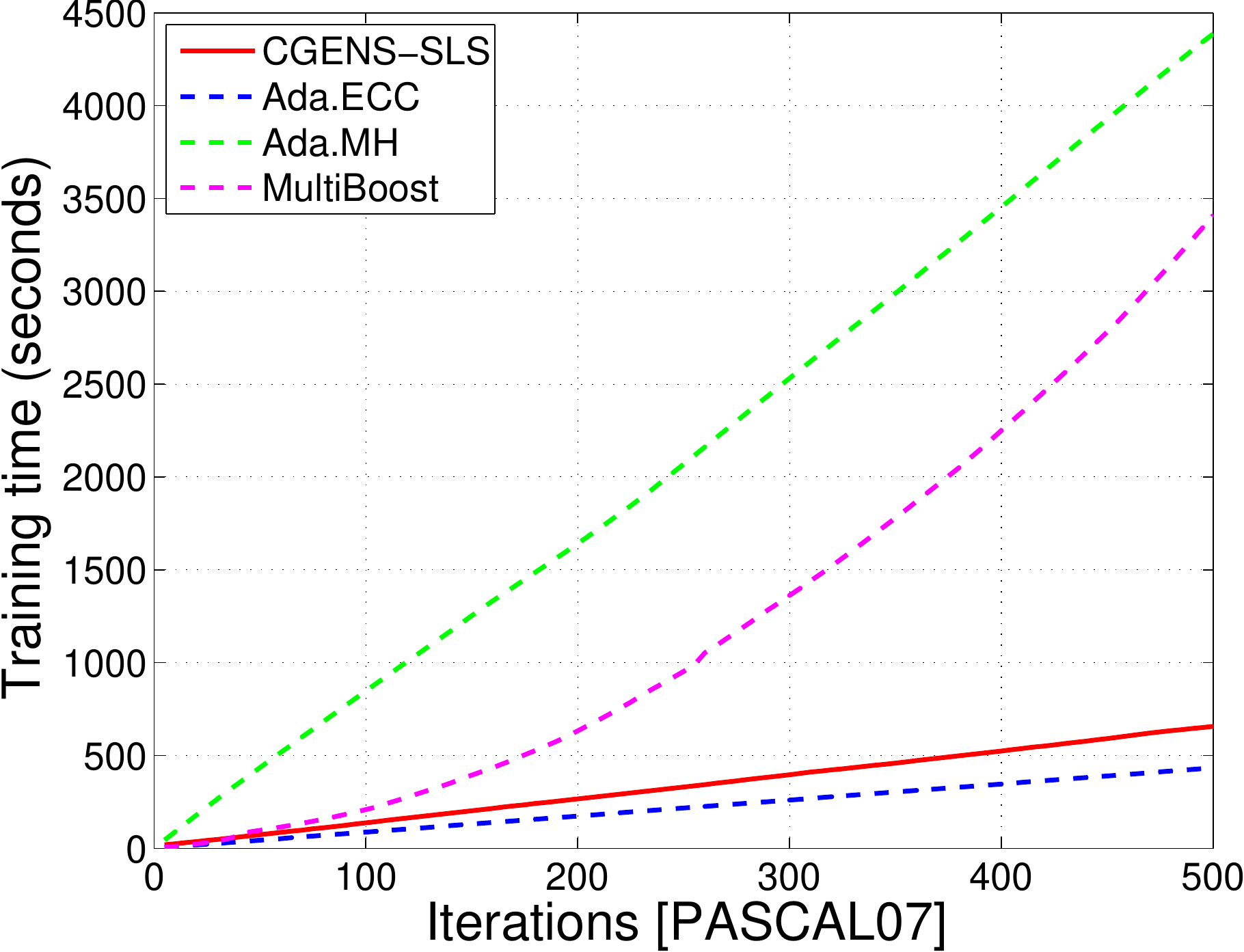}
    \includegraphics[width=0.186\textwidth, height=0.15\textwidth]{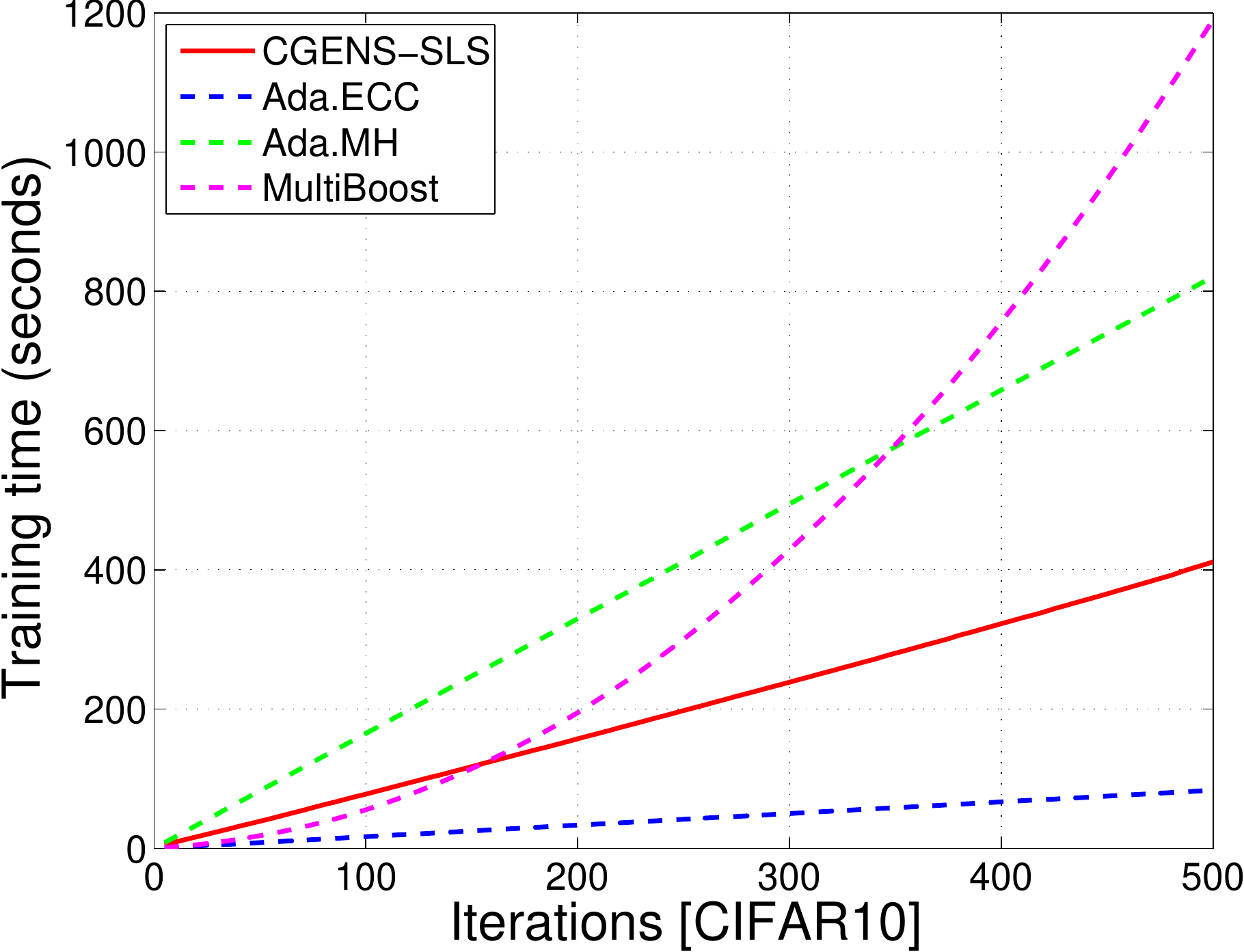}
    \includegraphics[width=0.186\textwidth, height=0.15\textwidth]{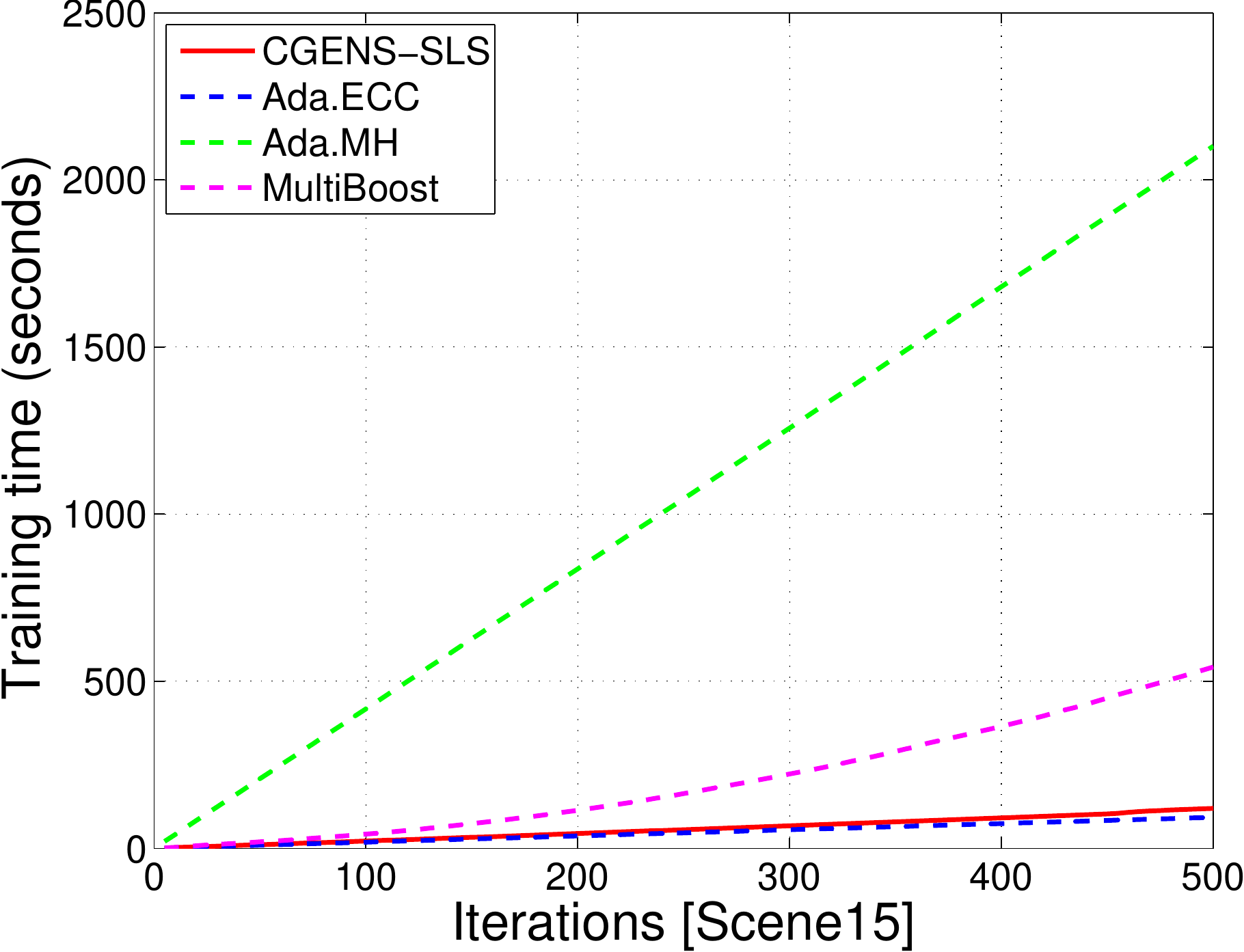}
    \includegraphics[width=0.186\textwidth, height=0.15\textwidth]{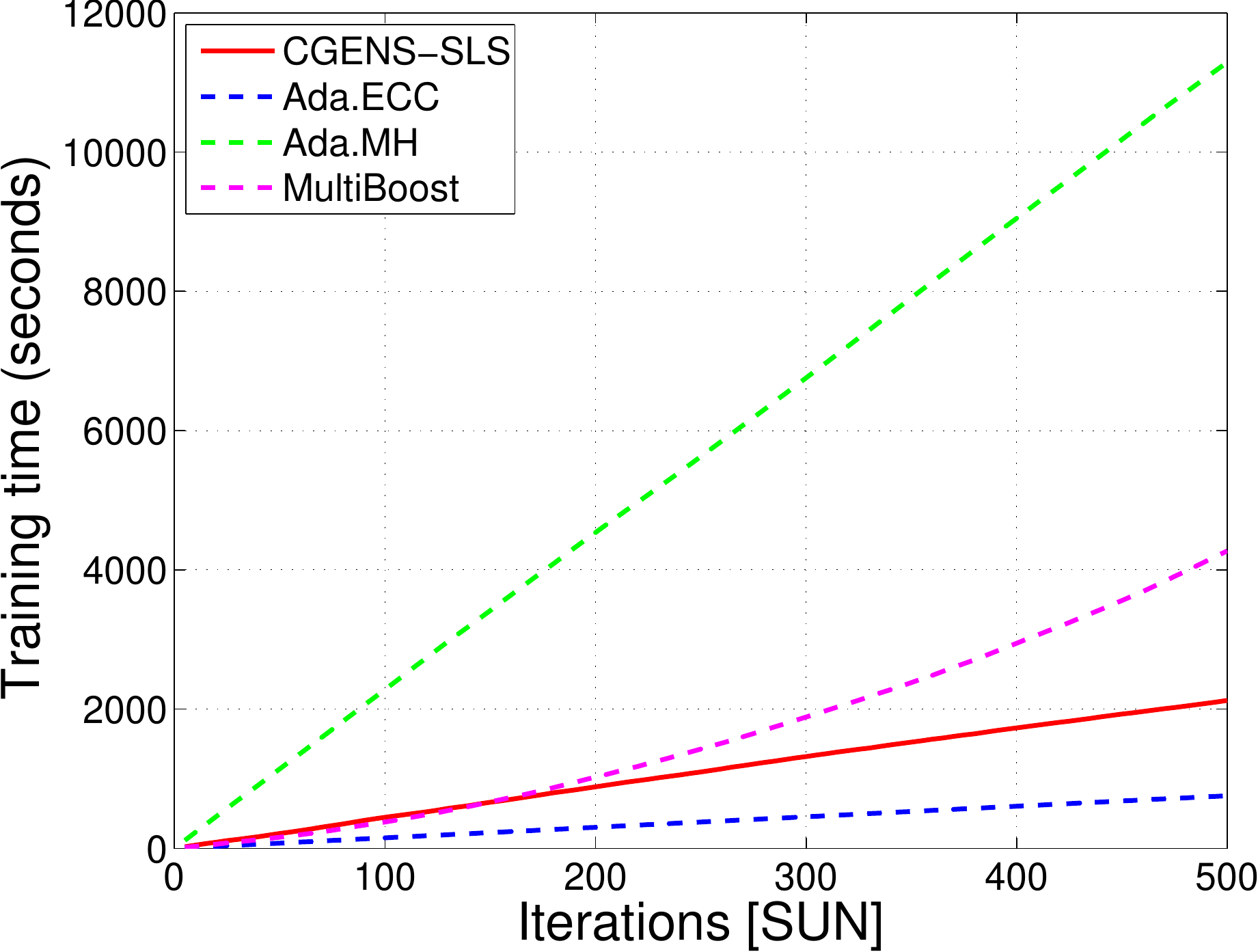}
    \vspace{.3cm}
    \caption{Test error (first row) and training time (second row) comparison with respect to different iterations on MNIST, Pascal07, CIFAR10, Scene15 and SUN datasets. Our \CGENS-SLS achieved overall best performance compared with AdaBoost.ECC, AdaBoost.MH and MultiBoost \cite{multiboost}.
    }
    \label{fig:mc-iter}
\end{figure*}

From Table \ref{table:lssvm}, we can see that the proposed \lssvm achieves overall best performance, especially on the vision datasets. Figure \ref{fig:mc-iter} shows the test error and training time comparison with respect to different iteration numbers on four image datasets. Our proposed \lssvm performs slightly better than all the other methods in terms of classification accuracy while being more efficient than AdaBoost.MH and MultiBoost.

    \section{Conclusion}
        Kernel methods are popular in domains even outside of the computer science community largely
        because they are easy to use and there are highly optimized software available.
        On the other hand, ensemble learning is being developed in a separate direction and has
        also found its applications in various domains. In this work, we show that
        {\em one can directly
         design ensemble learning methods from kernel methods like SVMs.
        }
        In other words, one may directly solve the optimization problems of kernel methods
        by using column generation technique.
        The learned ensemble model is equivalent to learning the explicit feature mapping functions
        of kernel methods. This new insight about the precise correspondence enables us to design
        new algorithms.
        In particular we have shown two examples of new ensemble learning methods which have roots in
        SVMs.
        Extensive experiments show the advantages of these new ensemble methods over conventional
        boosting methods in term of both classification accuracy and computation efficiency.

\section{Appendix}

\subsection{Solutions of the multi-class SLS-SVM and ensemble learning}

The Lagrangian of problem (\ref{EQ:SLSVM1}) is
    \begin{equation}
        \label{eq:slsvm:L}
    \begin{split}
        & L(\bw_1,\cdots,\bw_l,{\bf b},\bO,\bU)
        =
        \half \textstyle \sum_{\tau=1}^{l} \fnorm{ \bw_\tau} ^{2} +
        \\
        &\tfrac{C}{2} \textstyle \sum_{i=1}^{m} \textstyle \sum_{\tau=1}^{l}{\bO}_{i\tau}^2 -
        \textstyle \sum_{i=1}^{m} \sum_{\tau=1}^{l} \bU_{ij} \cdot
        \\
        &
        ( \bw_\tau^\T \Phi(\bx_i)+b_\tau-\bL_{i\tau}+\bO_{i_\tau}),
    \end{split}
    \end{equation}
where $\bU \in\mathbb{R}^{m\times l}$ is the collection of Lagrange variables corresponded to $\bO$.
The optimization problem can be solved by setting its first order
partial derivative with respect to the parameters $\bw_\tau, b_\tau,
\bL_{i\tau}$ and $\bO_{i\tau}$ to zeros:
    \begin{eqnarray}
    \begin{split}
         \frac{\partial L}{\partial \bw_\tau} &= \bw_\tau - \textstyle {\sum_{i=1}^{m}}\bU_{i\tau}\Phi(\bx_i) = \zeros,  \\%\Longrightarrow {\bf w}_j = HU_{:j}, \\
         \frac{\partial L}{\partial {b}_\tau} &= -\textstyle{\sum_{i=1}^{m}}\bU_{i\tau} = 0, \\
         \frac{\partial L}{\partial {\bO}_{i\tau}} &= C \bO_{i\tau} - \bU_{i\tau} = 0, \\ %
         \frac{\partial L}{\partial {\bU}_{i\tau}} &= \bw_\tau ^ \T \Phi( \bx_i ) + b_\tau - \bL_{i\tau} +\bO_{i\tau} = 0 .
    \end{split}
    \end{eqnarray}

    Use $\H\in\mathbb{R}^{J\times m}$ to denote the weak classifiers' response on the whole training data such that each column $\H_{:i}=\Phi(\bx_i)$. The previous conditions can be rewritten to the following form:
    \begin{subequations} \label{eq:lssvm:cond}
    \begin{align}
        \bw_\tau &= \H\bU_{:\tau},\label{eq:lssvm:cond:1}\\
         0 &= \bU_{:\tau} ^\T \ones, \label{eq:lssvm:cond:2}\\
        \bO_{:\tau} &= C^{-1}\bU_{:\tau}, \label{eq:lssvm:cond:3}\\
        \bL_{:\tau} &=  \H^\T \bw_\tau + b_\tau \ones  + \bO_{:\tau}  , \label{eq:lssvm:cond:4}
    \end{align}
    \end{subequations}
    where $\bO_{:\tau}$ and $\bL_{:\tau}$ is the $\tau$th column of $\bO$ and $\bL$, respectively.
    Substituting \eqref{eq:lssvm:cond:1} and \eqref{eq:lssvm:cond:3} into \eqref{eq:lssvm:cond:4}, we have
    \begin{eqnarray}
    \begin{split}
                        & \bL_{:\tau} = (\overbrace{\H ^\T \H +
                        C^{-1}\bI_m}^{\mbox{ def. as } \bS}) \bU_{:\tau} + b_\tau \ones  \\
        \Longrightarrow \ \ & \bS^{-1} \bL_{:\tau} = \bS^{-1} \bS \bU_{:\tau} + b_\tau \bS^{-1} \ones \\
        \Longrightarrow \ \ & \bS^{-1} \bL_{:\tau} = \bS^{-1} \bS \bU_{:\tau} + b_\tau \bS^{-1} \ones \\
        \Longrightarrow \ \ & \ones^\T \bS^{-1} \bL_{:\tau} = \overbrace{\ones^\T \bU_{:\tau}}^{=0,\mbox{ due to (\ref{eq:lssvm:cond:2})}} + b_\tau \ones^\T \bS^{-1} \ones \\
        \Longrightarrow \ \ & \begin{split}
                                & b_\tau = \frac{\ones^\T \bS^{-1} \bL_{:\tau}}{\ones^\T \bS^{-1} \ones},
                                & \bU_{:\tau} = \bS^{-1}(\bL_{:\tau}-b_\tau \ones)
                              \end{split} \\
        \Longrightarrow \ \ & \begin{split}
                                & \bb = (\frac{\ones^\T \bS^{-1} \bL}{\ones^\T \bS^{-1} \ones})^\T,
                                & \bU = \bS^{-1}(\bL - \ones \bb^\T).
                              \end{split}
    \end{split}
    \end{eqnarray}

    The inverse for $\bS$ can be computed efficiently as follows.
    Suppose $\H_{(J)},\bS_{(J)},\bS^{-1}_{(J)}$ and $\H_{(J+1)},\bS_{(J+1)},\bS^{-1}_{(J+1)}$ are matrices in the $J$th and $(J+1)$th iteration, respectively.
    We have $\H_{(J+1)}=\left[ \H_{(J)}^{\T}\;{\bf h}_{J+1}\right]^{\T}$, where ${\bf h}_{J+1}=\left[ \wl_{J+1}({\bf x}_1),\wl_{J+1}({\bf x}_2),\cdots,\wl_{J+1}({\bf x}_n)\right]^{\T}$.
    It is easy to see that $\bS_{(J)},\bS^{-1}_{(J)},\bS_{(J+1)}$ and $\bS^{-1}_{(J+1)}$ are symmetric matrices. So,
    \begin{eqnarray*}
    \begin{split}
        \bS^{-1}_{(J+1)}  =& \left(C^{-1}\bI_m + \H_{(J+1)}^{\T}\H_{(J+1)}\right)^{-1}\\
                        =& \left(C^{-1} \bI_m + \H_{(J)}^{\T} \H_{(J)}+{\bf h}_{J+1}{\bf h}_{J+1}^{\T}\right)^{-1} \\
                        =& \left(\bS_{(J)} +{\bf h}_{J+1}{\bf h}_{J+1}^{\T}\right)^{-1} \\
                        =& \bS^{-1}_{(J)}-\bS^{-1}_{(J)}{\bf h}_{J+1}\left(1+{\bf h}_{J+1}^{\T}\bS^{-1}_{(J)}{\bf h}_{J+1}\right)^{-1}{\bf h}_{J+1}^{\T}\bS^{-1}_{(J)}.
    \end{split}
    \end{eqnarray*}
    Let $\bs_{J+1} = \bS^{-1}_{(J)}\bh_{J+1}$, the update process finally is
    \begin{equation}
        \bS^{-1}_{(J+1)} = \bS^{-1}_{(J)}-\frac{\bs_{J+1} \bs_{J+1}^{\T}}{1+\bh_{J+1}^{\T}\bs_{J+1}}.
    \end{equation}

    \begin{figure*}
    \begin{center}
    {\includegraphics[width=0.19\linewidth]{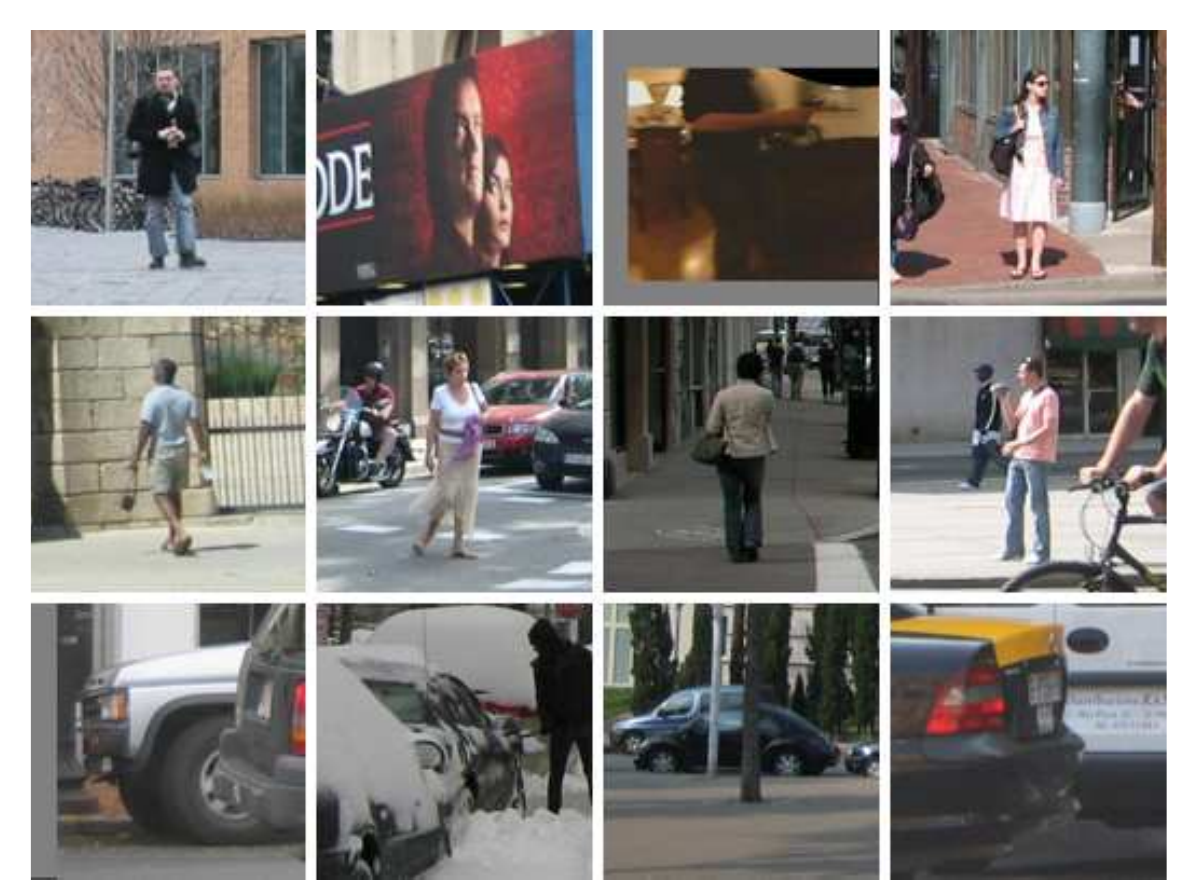}}
    {\includegraphics[width=0.19\linewidth]{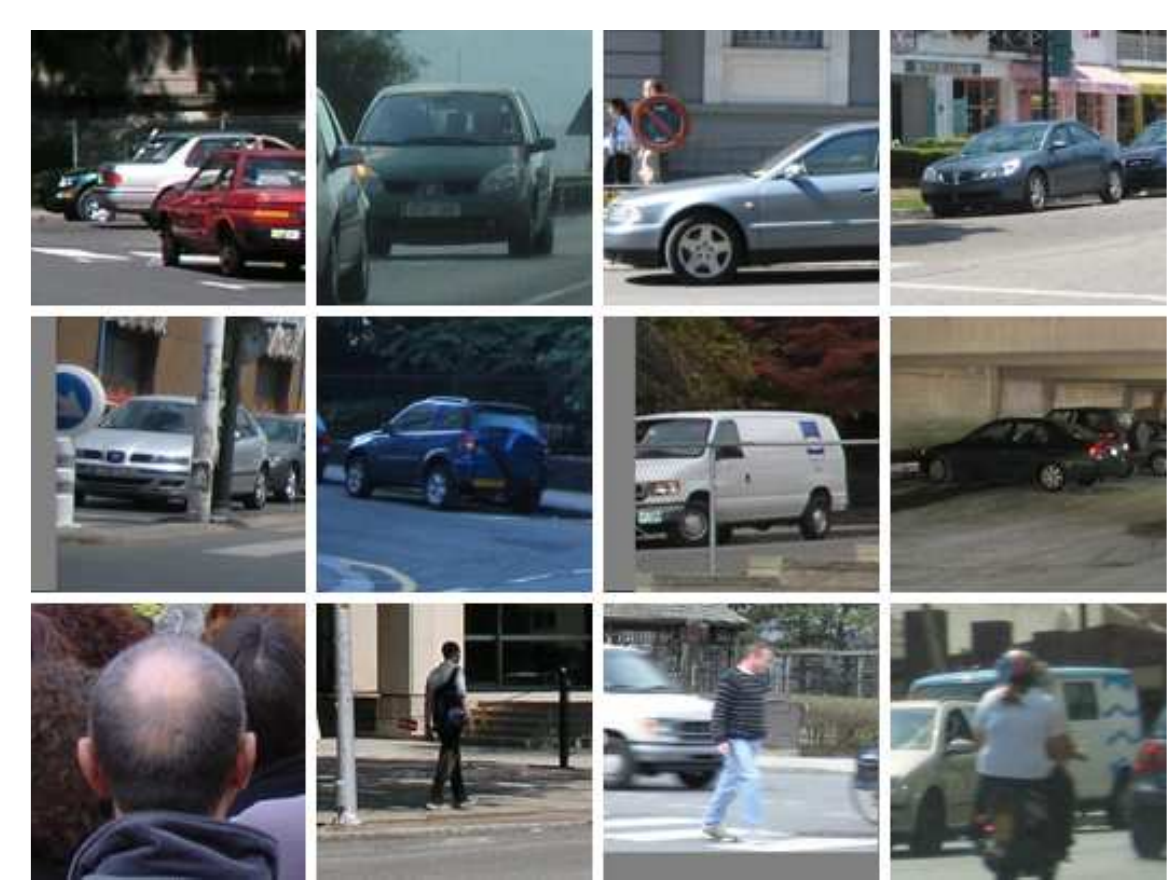}}
    {\includegraphics[width=0.19\linewidth]{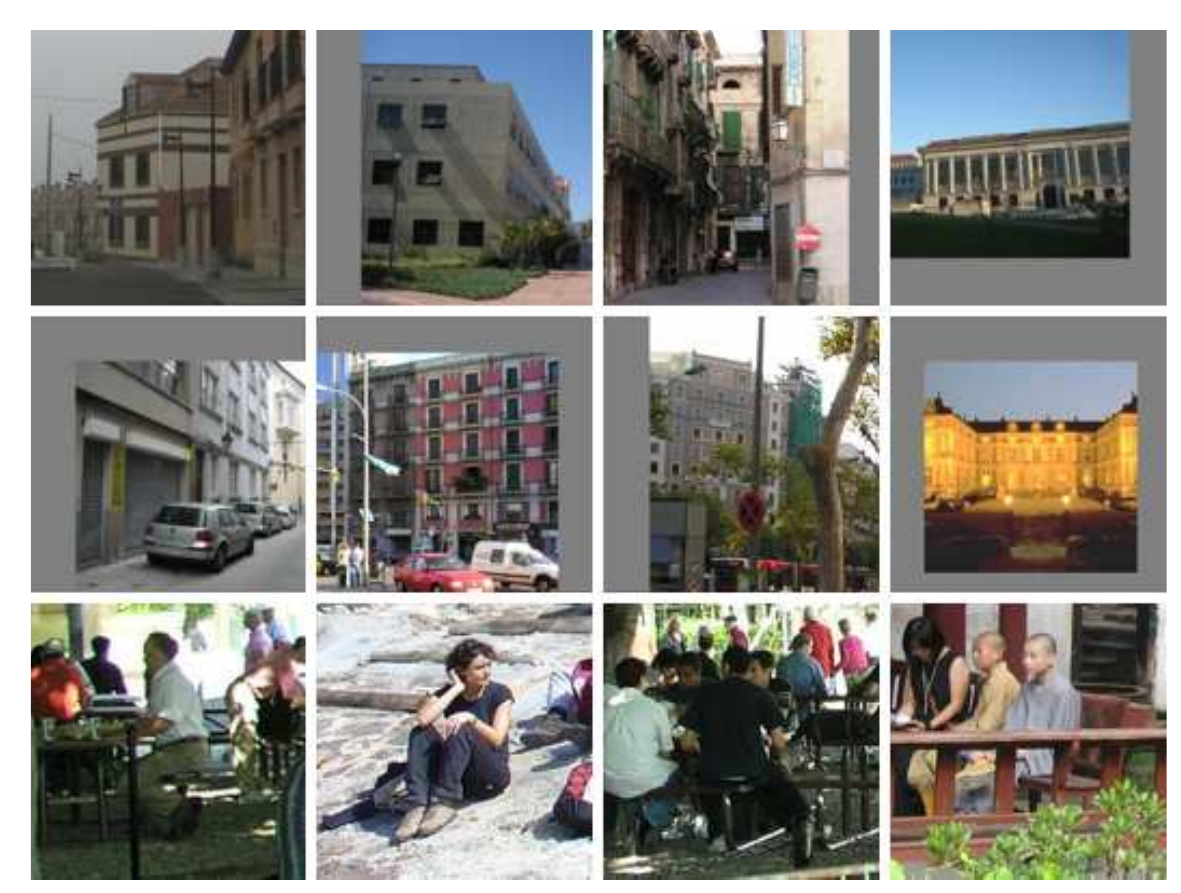}}
    {\includegraphics[width=0.19\linewidth]{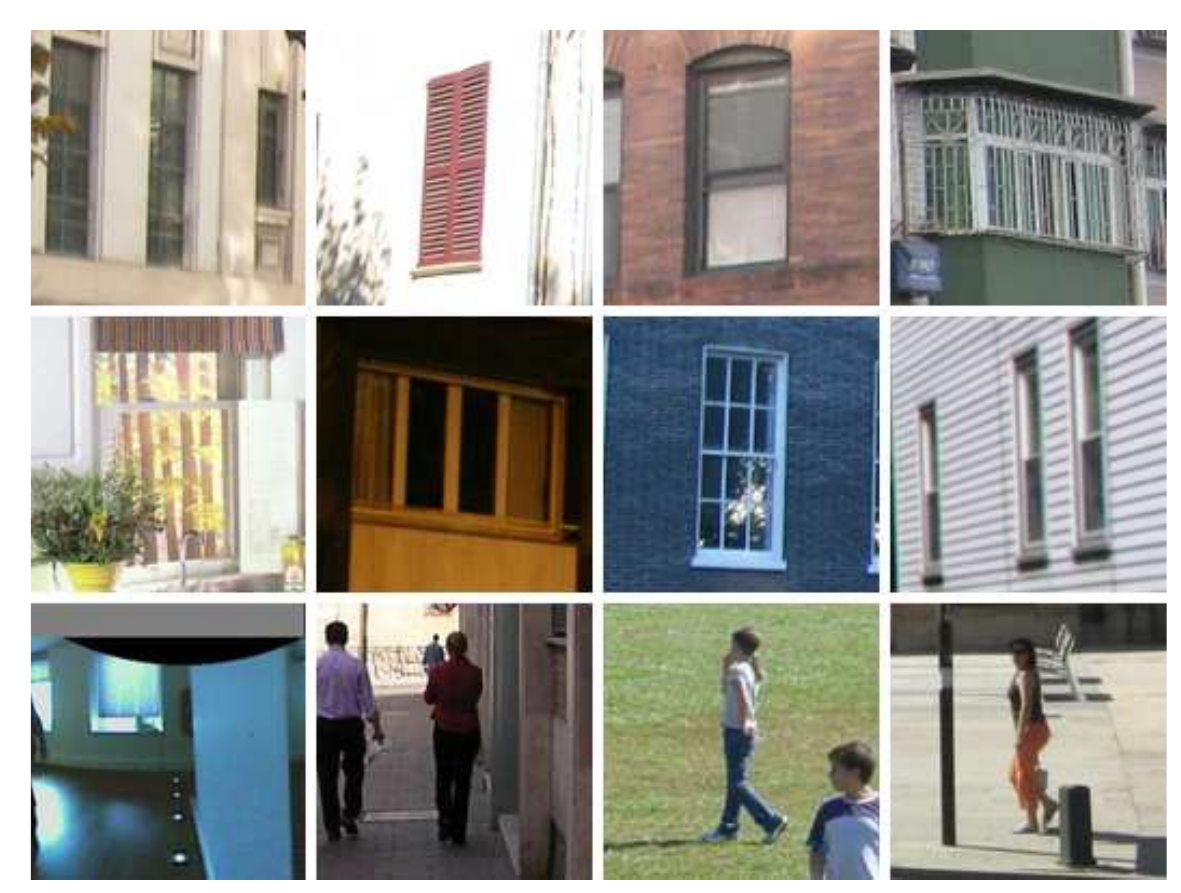}}
    {\includegraphics[width=0.19\linewidth]{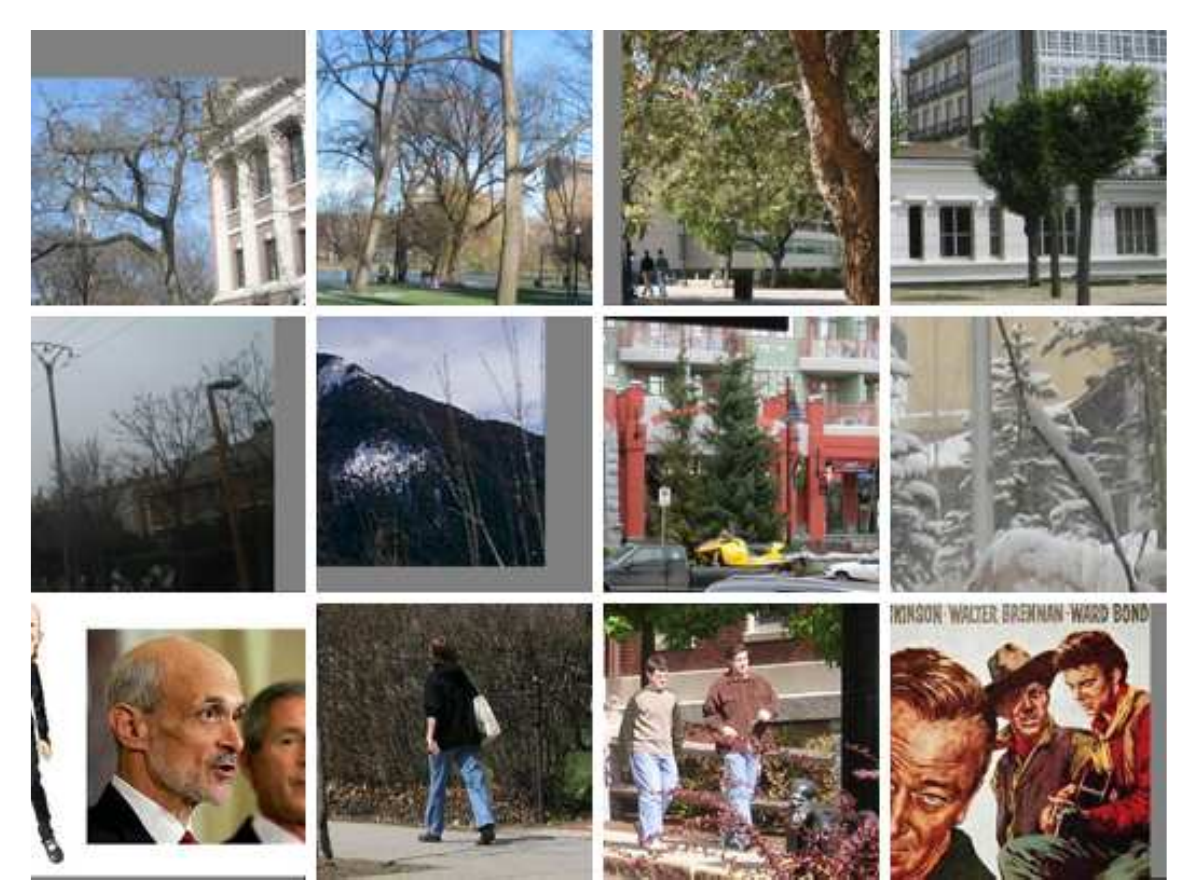}}
    \end{center}
    \vspace{-.3cm}
    \caption{Some examples of correctly classified (top two rows) and misclassified (bottom row) images by \lssvm in LabelMe data set.}
    \label{fig:example_lssvm}
    \end{figure*}

\subsection{Binary classification on Spambase data set}

\begin{table}
    \centering
    \begin{tabular}{l| l}
    \hline
    \textbf{Algorithm} &  \textbf{Error rate (\%)}  \\
    \hline
    \hline
    AdaBoost & 6.32$\pm$0.66  \\
    Ours & 5.80$\pm$0.46  \\
    \hline
    \end{tabular}
        \caption{Error rates (\%) of AdaBoost and the proposed method on spam dataset using decision stump.}
\label{tabSpam}
\end{table}

\begin{figure*}     %
    \begin{center}
    \includegraphics[width=0.80\textwidth]{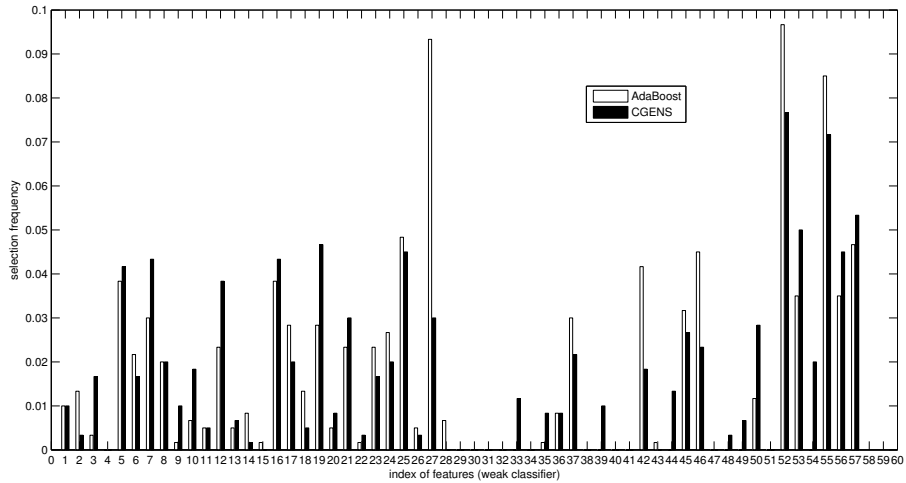}
    \end{center}
    \caption{The frequency of different features being selected on the \textbf{spam} dataset.}
        \label{figSpam}
\end{figure*}

We performed experiments on the UCI Spam dataset to demonstrate the feature selection of the proposed method when using decision stump as weak learner.
The task is to differentiate spam emails according to word frequencies.
We use AdaBoost as a baseline.
The maximum iterations are both set to 60 due to fast convergence and no overfitting observed hereafter.
We use 5-fold cross validation to choose the best hyper parameter $C$ in \CGENS.
The results, shown in Table \ref{tabSpam}, are reported over 20 different runs with training/testing ratio being $3:2$.
Fig. \ref{figSpam} plots the average frequencies over the $20$ rounds.
As can be observed, both algorithms select important features such as ``free" (feature \#16),
``hp" (25) and ``!" (52) with high frequencies. As for the other features, the two methods showed different inclinations.
\CGENS tends to select features like ``remove" (7), ``you" (19), ``\$" (53) which are intuitively meaningful for the classification.
On the contrary, the favorite ones of AdaBoost are ``george" (27), ``meeting" (42) and ``edu" (46) which are more irrelevant for spam email detection.
This explains why our method slightly outperformed AdaBoost in test accuracy.

{

}

\end{document}